\documentclass{article}

 \usepackage[main, final]{neurips_2026}
\usepackage[utf8]{inputenc} % allow utf-8 input
\usepackage[T1]{fontenc}    % use 8-bit T1 fonts
\usepackage{hyperref}       % hyperlinks
\usepackage{url}            % simple URL typesetting
\usepackage{booktabs}       % professional-quality tables
\usepackage{amsfonts}       % blackboard math symbols
\usepackage{nicefrac}       % compact symbols for 1/2, etc.
\usepackage{microtype}      % microtypography
\usepackage{xcolor}         % colors

\usepackage{amsmath,amsfonts,bm}

\def\eps{{\epsilon}}

\def\ra{{\textnormal{a}}}

\def\rw{{\textnormal{w}}}

\DeclareMathAlphabet{\mathsfit}{\encodingdefault}{\sfdefault}{m}{sl}
\SetMathAlphabet{\mathsfit}{bold}{\encodingdefault}{\sfdefault}{bx}{n}

\DeclareMathOperator{\sign}{sign}

\usepackage{microtype}
\usepackage{graphicx}
\usepackage{subcaption}
\usepackage{hyperref}

\usepackage{amsmath}
\usepackage{amssymb}
\usepackage{mathtools}
\usepackage{amsthm}

\usepackage{url}
\usepackage{comment}

\definecolor{myblue}{rgb}{0.1,0.2,0.75}

\hypersetup{ %
pdftitle={},
pdfauthor={},
pdfsubject={},
pdfkeywords={},
pdfborder=0 0 0,
pdfpagemode=UseNone,
colorlinks=true,
linkcolor=myblue,
citecolor=myblue,
filecolor=myblue,
urlcolor=myblue,    
pdfview=FitH}

\usepackage{titletoc}

\newcommand\DoToC{%
  \startcontents
  \printcontents{}{1}{\textbf{Table of Contents}\vskip3pt\hrule\vskip5pt}
  \vskip3pt\hrule\vskip5pt
}

\theoremstyle{plain}
\newtheorem{theorem}{Theorem}[section]

\newtheorem{proposition}[theorem]{Proposition}
\newtheorem{lemma}[theorem]{Lemma}
\theoremstyle{definition}

\newtheorem{remark}[theorem]{Remark}

\newtheorem{observation}[theorem]{Observation}
\newcommand{\Be}{\begin{equation}}
	\newcommand{\Ee}{\end{equation}}

\newcommand{\R}{\mathbb{R}}

\renewcommand{\S}{\mathbb{S}}
\newcommand{\be}{\begin{equation}}
\newcommand{\ee}{\end{equation}}

\newcommand{\Bes}{\begin{eqnarray*}}
\newcommand{\Ees}{\end{eqnarray*}}

\newcommand{\Bs}{\begin{split}}

\def\beq{\begin{equation}}
\def\eeq{\end{equation}}
\def\bb1{{1\!\!1}}

\def\triangle{\Delta}
\def\bega{\begin{aligned}}
\def\enda{\end{aligned}}
\def\R{\mathbb{R}^2}

\def\lw{\left}
\def\rw{\right}

\def\R{\mathbb{R}}
\def\wtd{\widetilde}
\def\la{\langle}
\def\ra{\rangle}

\def\bcase{\begin{cases}}
\def\ecase{\end{cases}}

\def\bmx{\begin{bmatrix}}
\def\emx{\end{bmatrix}}

\def\bitem{\begin{itemize}}
\def\eitem{\end{itemize}}

\title{First-Token Attraction in Mamba Dynamics}

\author{%
  \parbox[t]{0.96\textwidth}{\centering\normalfont
    \textbf{Trinh T. Nguyen}$^{1}$\thanks{Co-first authors}\quad
    \textbf{Minh-Khoi Nguyen-Nhat}$^{2}$\footnotemark[1]\quad
    \textbf{Duy-Tung Pham}$^{3}$\\[2pt]
    \textbf{Hoang-Son Do}$^{4,5}$\quad
    \textbf{Tan Minh Nguyen}$^{2,\dagger}$\quad
    \textbf{Thieu N. Vo}$^{6,\dagger}$\\[4pt]
    {\small
    $^{1}$ School of Mathematical Sciences, Shanghai Jiao Tong University, China\\
    $^{2}$ Department of Mathematics, National University of Singapore\\
    $^{3}$ Khoury College of Computer Sciences, Northeastern University, USA\\
    $^{4}$ Institute of Mathematics, Vietnam Academy of Science and Technology\\
    $^{5}$ University of Science and Technology of Hanoi, Vietnam Academy of Science and Technology\\
    $^{6}$ Department of Computer Science, University of Bath, United Kingdom}\\[3pt]
    {\footnotesize
    $^{1}$\texttt{q0004349@sjtu.edu.cn}\quad
    $^{2}$\texttt{khoi.nguyen@u.nus.edu}\quad
    $^{3}$\texttt{pham.duy@northeastern.edu}\\
    $^{4}$\texttt{dhson@math.ac.vn}\quad
    $^{5}$\texttt{do-hoang.son@usth.edu.vn}\\
    $^{2}$\texttt{tanmn@nus.edu.sg}\quad
    $^{6}$\texttt{ntv22@bath.ac.uk}}
  }
}

\begin{document}

\maketitle
{\def\thefootnote{$\dagger$}\footnotetext{Co-last authors. Correspondence to: \texttt{q0004349@sjtu.edu.cn} \& \texttt{ntv22@bath.ac.uk}.}}

\begin{abstract}
  We study token dynamics in deep Mamba models, with a particular focus on the attracting role of the first token. Unlike attention mechanisms, where token interactions are governed by positive scalar coefficients, Mamba dynamics involve full matrix-valued interaction coefficients with no positivity or definiteness guarantees. This structural difference makes Mamba dynamics substantially harder to analyze and prevents the direct application of existing techniques for attention-based models. To overcome this challenge, we introduce a new analytical framework based on a high-dimensional invariant box and a sharp exit-time argument, which together recover an effective form of sign-definiteness. Using this framework, we characterize the possible limit points of token trajectories and prove explicit exponential convergence rates. Our analysis reveals that the first token has privileged dynamics: its limiting direction attracts subsequent tokens, inducing token clustering and attention-sink phenomena in deep Mamba models. We further show that this attraction mechanism extends to broader sequence dynamics, including state-space models, causal attention, and classical time-series models. Experiments on Falcon-Mamba-7B validate our predictions, showing that concentration on the first token increases with depth and dominates in later layers.
\end{abstract}

%\vspace{-0.2cm}
\section{Introduction}
%\vspace{-0.2cm}

Modern sequence modeling has been dominated by transformers, while state space models (SSMs) have emerged as an important alternative. Transformers have set the standard for representation learning, yet their quadratic complexity with respect to sequence length poses challenges for efficiency and scalability \citep{vaswani2017attention}. In parallel, SSMs have reemerged as powerful alternatives that combine principles of recurrence and convolution with ideas from classical control theory \citep{kalman1960new}. Recent innovations in parameterization and implementation have made SSMs substantially more efficient \citep{gu2022efficiently, gu2021combining, gu2022parameterization, gupta2022diagonal, li2023makes, kosma2023time, orvieto2023resurrecting, smith2023simplified}, enabling them to capture long-range dependencies effectively \citep{gu2020hippo,nguyen2026muonssm} and to excel in benchmark tasks such as the Long Range Arena \citep{tay2021long}. Beyond efficiency, these models have demonstrated strong empirical performance in domains as varied as audio, speech, and vision \citep{goel2022s, nguyen2022s4nd, saon2023diagonal}.

Within this line of development, the Mamba architecture has emerged as a selective SSM with striking empirical results. Its distinctive component, the selective state-space ($\operatorname{S6}$) layer, introduces input-dependent parameters that grant the model content awareness \citep{gu2024mamba}. This design has proven effective across a wide spectrum of applications, including language modeling \citep{pioro2024moe, lenz2025jamba}, vision and video analysis \citep{liu2024generalization, zhu2024vision, yang2024vivim, li2024videomamba}, medical imaging \citep{ma2024semi, wang2024mamba}, tabular data \citep{ahamed2024mambatab}, reinforcement learning \citep{ota2024decision}, 3D perception \citep{liang2025pointmamba}, graph learning \citep{wang2024graph}, and even $N$-dimensional sequence modeling \citep{li2024mamba}. The breadth of these successes has been linked to Mamba’s ability to generate highly adaptive and expressive representations across modalities \citep{jafari2025mambalrp, patro2024mamba, xu2024survey, zhu2024vision}.

The study of token dynamics--the evolution of token representations within a model--has proven to be a powerful framework for understanding transformers, offering insights into equilibrium structures, stability properties, and clustering phenomena. 
In contrast, the token dynamics of selective SSMs have received little attention, despite the strong empirical performance of architectures such as Mamba. 
% Existing analyses rely on restrictive assumptions, such as one-dimensional embeddings or the exclusion of normalization layers, which substantially oversimplify the behavior of practical implementations \cite{vo2024demystifying}. 
% Real-world Mamba models operate in high-dimensional embedding spaces and incorporate layer normalization, leading to internal dynamics that are considerably richer than current theories capture. 
% As these models are scaled and deployed more broadly, the absence of a rigorous theoretical foundation for their token dynamics not only leaves fundamental questions unanswered but also hinders a principled explanation of the mechanisms that drive their empirical success.

%\vspace{-0.2cm}
\subsection{Our Contributions}
%\vspace{-0.2cm}

In this paper, we investigate the dynamical properties of tokens in deep pretrained Mamba models.
% under more realistic and practically relevant assumptions on model parameters. 
Prior theoretical studies often rely on restrictive conditions such as the absence of normalization and one-dimensional token embeddings (see \citep{vo2024demystifying}). 
We consider the continuous-time limit of pretrained Mamba models without these restrictive assumptions
and systematically analyze the asymptotic behavior of token trajectories over time. 
Our main contributions are summarized as follows:

\begin{itemize}
    
    % \item We derive a dynamical system governing the continuous-time limit of pretrained deep Mamba models. In contrast to previous studies, our analysis incorporates layer normalization and investigates token dynamics on high-dimensional spheres. %In addition, we allow the parameters to be time-dependent.

    \item We construct a new theoretical framework based on a high-dimensional invariant box for the Mamba dynamics and establish a sharp exit-time argument to determine when token trajectories remain confined to this domain.
    This framework overcomes the difficulties arising in previous approaches for attention settings.

    \item We characterize the set of limit points of token trajectories in Mamba models and establish explicit exponential convergence rates (Theorems~\ref{thm:first_token_main},~\ref{thm:first_token_D=2_main},~\ref{thm:all-conv_main} and~\ref{thm:nonautonomous2_main}. %Proposition~\ref{prop:positive}). 
    Our theoretical analysis further demonstrates that the first token exerts an attracting influence on the remaining tokens, a phenomenon that we empirically validate.
    
    % \item We extend our theoretical analysis to a broader class of autoregressive sequence models, including state space models, causal Transformers, and classical time-series models (Theorem~\ref{thm:MIT_main}). We then strengthen a result established in \citep{karagodin2024clustering} about the asymptotic convergence of tokens in causal attention dynamic.
    
    \item We demonstrate a practical implication of our theoretical results to offer a principled explanation of the attention sink effect in Mamba models via the attraction of the first token.
    % We demonstrate the practical implications of our theoretical results through two applications: (i) offering a principled explanation of the attention sink effect in Mamba models via the attracting role of the first token, and (ii) randomizing token order during training to diversify clustering.
  
    % \item Novel Mathematical Contribution:
    % Our analysis provides a new and broadly applicable framework for understanding token dynamics in state-space (Mamba) architectures. While inspired by the recent Transformer analysis of Karagodin et al.~(2024), the mathematical structure of Mamba is fundamentally different: the effective interaction coefficients are \emph{not sign-definite}, so none of the positivity-based techniques used for attention apply. This sign-instability is not a technical nuisance—it is a core structural obstruction that arises in practical Mamba implementations. To address this, we construct a high-dimensional \emph{invariant box} that captures all admissible trajectories of the nonlinear coefficient system, and we prove—via a sharp \emph{exit-time argument}—that the coefficients of a deep Mamba layer remain inside this region for all depths. This allows us to show that Mamba \emph{recovers sign-definiteness dynamically}, unlocking PDE-style tools such as quantitative Grönwall estimates to derive \emph{explicit exponential convergence rates}. 

\end{itemize}

To validate our theoretical findings and their applications, we conduct extensive experiments by visualizing attention concentration in Falcon-Mamba-7B across different depths. %and by testing the impact of random token reordering in MambaVision-T on ImageNet-1K classification benchmarks.
These experimental results confirm the predictive accuracy of our analysis and underscore its practical benefits in real-world scenarios.

%\vspace{-0.2cm}
\paragraph{Mamba dynamics is far more complicated than causal attention dynamics.}
Indeed, while our work is inspired by the recent analysis of Transformer dynamics in \cite{karagodin2024clustering}, the mathematical structure of Mamba is much more complex. In the attention setting, the effective interaction coefficients are positive scalars as they are normalized via softmax. The positivity of these interaction coefficients ensures sign-definiteness and enables the use of powerful tools from monotone and cooperative dynamical systems. In contrast, for Mamba, the effective interaction coefficients are full matrices with no positivity or definiteness guarantees. This lack of sign-definiteness is not a minor technical inconvenience; it is an intrinsic structural feature of practical Mamba architectures and invalidates essentially all positivity-based techniques that underpin the current theory of attention dynamics. As a consequence, techniques used for analyzing causal attention dynamics, including hemisphere-based Lyapunov methods~\cite{abellaconsensus} and Shub-type genericity arguments~\cite{karagodin2024clustering}, completely fail in this architecture.

%\vspace{-0.2cm}
\paragraph{Our new analytical framework.}
To overcome the above challenge, we develop a new analytical framework based on a high-dimensional invariant box for the dynamical system governing the continuous-time limit of Mamba, together with a sharp exit-time argument that recovers effective sign-definiteness and determines when token trajectories remain confined to this domain. Within this framework, we characterize all possible token limit points and derive explicit exponential convergence rates toward them. This analysis yields a theoretical explanation for a widely observed empirical phenomenon: the limit point of the first token exerts an attracting influence on subsequent tokens, thereby inducing clustering and attention-sink phenomena. Finally, we extend our analysis of this attracting effect to a broad class of autoregressive sequence models, including state-space models, causal Transformers, and classical time-series models.

% We overcome the above difficulty by constructing a new framework based on a high-dimensional invariant box for the Mamba dynamic and establishing a sharp exit-time argument to recover effective sign-definiteness.

% To overcome this challenge, we develop a new analytical framework that constructs a geometric invariant domain for the dynamical system governing the continuous-time limit of Mamba and employs a sharp exit-time argument to determine when token trajectories remain confined to this domain. 
%     Within this framework, we characterize all possible token limit points and derive explicit exponential convergence rates toward them. 
%     Our theory explains a widely observed empirical phenomenon: the limit point of the first token exerts an attracting influence on subsequent tokens, inducing clustering and attention-sink phenomena. 
%     Finally, we extend our analysis of this attracting effect to a broad class of autoregressive sequence models, including state-space models, causal Transformers, and classical time-series models.
% {\color{red} ...can we say more about this???...}

% \section{Related Works}

\textbf{Organization.} Section~\ref{sec:mamba_dynamic} derives a dynamical system over the unit sphere arising from the continuous-time limit of Mamba. Section~\ref{sec:theoretical-justifications} analyzes the limitations of the first token and proves its attraction effect on subsequent tokens. Section~\ref{sec:simulation} validates the theory in practical Mamba settings, while Section~\ref{sec:experiment} applies the results to predict attention-sink phenomena in Mamba models.

 %\vspace{-0.4cm}
\section{Mamba Dynamics as a Flow Map on Sphere}\label{sec:mamba_dynamic}
 %\vspace{-0.2cm}

\subsection{Background -- $\operatorname{S6}$ layer}
 %\vspace{-0.2cm}

The building block of a selective state-space (Mamba) model is a Mamba block (see \citep{dao2024transformers}). At the heart of the Mamba block is the $\operatorname{S6}$ layer, which plays a key role in the success of deep Mamba models. 
% In order to conduct a thoughtful study on the dynamical properties of tokens in a deep state space model, we will consider a simplified setting of the Mamba block, which contain the S6 and the layer normalization, with a skip connection.

 %\vspace{-0.2cm}
\paragraph{$\operatorname{S6}$ layer.}
Formally, an $\operatorname{S6}$ layer is defined as a function that maps a sequence of tokens $\mathbf{x} = (x_1, \ldots, x_L) 
	 \in \mathbb{R}^{D \times L}$ to another sequence of tokens $\mathbf{y} = (y_1, \ldots, y_L) 
	 \in \mathbb{R}^{D \times L}$ (with the same number of channels $D$).
In this context, each vector $x_l \in \mathbb{R}^D$ (or $y_l \in \mathbb{R}^D$) represents a token/word, while the whole sequence $\mathbf{x} = (x_1, \ldots, x_L)$ (or $\mathbf{y} = (y_1, \ldots, y_L)$) is called a prompt/sentence.

For each $d=1, \ldots, D$, the $d$-th channel output $\hat{y}_d = (y_{d1}, \ldots, y_{dL}) \in \mathbb{R}^{L}$ is determined recurrently from the corresponding $d$-th channel input $\hat{x}_d = (x_{d1}, \ldots, x_{dL}) \in \mathbb{R}^{L}$ via a sequence of hidden states $h_{d1}, \ldots, h_{dL} \in \mathbb{R}^{N}$ as follows:
	\begin{equation}\label{eq:recurrent_rule}
		\left\{
			\begin{aligned}
				&h_{dl} = \overline{A}_{dl} \cdot h_{d,l-1} + \overline{B}_{dl} \cdot x_{dl}, \quad h_{d0} = 0, \\
				&y_{dl} = C_l \cdot h_{dl},
			\end{aligned}
		\right.
	\end{equation}
for $l = 1, \ldots, L$.
Here, the matrices $\overline{A}_{dl}$, $\overline{B}_{dl}$, and $C_l$ are input-dependent and time-varying, determined by
\begin{align}\label{eq:recurrent_ABC}
	\overline{A}_{dl} = e^{-a_d \Delta_d(x_l)}, \quad
    \overline{B}_{dl} = \Delta_d(x_l) S_B \cdot x_l, \quad \text{and} \quad
    C_l = (S_C \cdot x_l)^{\top},
\end{align}
where $\Delta_d \,:\, \mathbb{R}^D \to \mathbb{R}$ is the step size function at the $d$-th channel and is defined for each $u \in \mathbb{R}^D$ by 
\begin{align}
    \Delta_d(u) = \text{softplus}(S_{\Delta,d} \cdot u) = \ln \left( 1+e^{S_{\Delta, d} \cdot u} \right).
\end{align}
The hidden positive coefficients $a_d \in \mathbb{R}_{>0}$, and the step size vectors $S_{\Delta,d} \in \mathbb{R}^{D}$ for $d = 1, \ldots, D$, as well as the input and output matrices $S_B, S_C \in \mathbb{R}^{N \times D}$, are learnable from input data. 
% The matrices $A_d$ are diagonal matrices with negative eigenvalues. 
% In recent developments, the matrices $A_d$ are chosen to be of the scalar form $A_d = -a_d I_N$ for some positive number $a_d$ (see \citep{dao2024transformers}), which does not compromise model performance. We will also use this scalar form for $A_d$ in our paper.
% In our context, the matrix $S_C^{\top} S_B \in \mathbb{R}^{D \times D}$, which we will refer to as the \emph{input-output matrix}, plays an essential role in characterizing the dynamical properties of tokens.

By eliminating the hidden states $h_{dl}$ in system~\eqref{eq:recurrent_rule} and recall that $y_l=(y_{1l},\ldots,y_{DL})$, we obtain the following form 
(see also \citep{ali2024hidden}):
% \begin{equation}\label{eq:convolution_rule_1}
%     y_{dl} = \sum\limits_{j=1}^l A_{d,lj}(\mathbf{x}) x_{dj},
% \end{equation}
% or equivalently,
\begin{equation}\label{eq:convolution_rule_D}
    y_{l} = \sum\limits_{j=1}^l A_{lj}(\mathbf{x}) x_{j},
\end{equation}
where $A_{lj} = \operatorname{diag}(A_{1,lj},\ldots,A_{D,lj})$ and $S_{BC}=S_C^{\top} S_B$ as well as
\begin{equation}\label{eq:hidden-attn}
    A_{d,lj}(\mathbf{x}) = \left\{
    \begin{aligned}
        & x_l^{\top} S_{BC} x_l \cdot \Delta_d(x_l), &\text{if } l = j,\\
        &x_l^{\top} S_{BC} x_j \cdot \Delta_d(x_j) \cdot \exp \Big( -a_d \sum\limits_{k=j+1}^l \Delta_d(x_k) \Big), &\text{if } l > j.
    \end{aligned}
    \right.
\end{equation}
% In the above formulation, the upper triangular matrices $P_d$ ($d=1, \ldots, D$) represent the hidden attention scores between the $d$-th channel input and output. Following the setting of \citep{ali2024hidden}, we call $P = [P_1, \ldots, P_D]$ the hidden attention tensor of the $\operatorname{S6}$ layer.

In our context, the matrix $S_{BC}=S_C^{\top} S_B \in \mathbb{R}^{D \times D}$, which we will refer to as the \emph{input-output matrix}, plays an essential role in characterizing the dynamical properties of tokens.

\subsection{Mamba Dynamic as Flow Map on Sphere}

All practical implementations of Mamba models incorporate normalization~\citep{ba2016layer}, most commonly realized as root mean square (RMS) normalization~\citep{zhang2019root}. In this approach, the sequence of tokens at each layer is normalized by dividing each token by its Euclidean norm. This operation enhances stability by preventing divergence, thereby reducing rounding errors and avoiding numerical overflow. Consequently, the resulting dynamics can be viewed as evolving on the unit sphere $\mathbb{S}^{D-1}$ throughout.

Follow the framework of \citep{karagodin2024clustering}, a Mamba model including $\operatorname{S6}$ layer and RMS normalization with skip connection can then be viewed as a flow map on $(\mathbb{S}^{D-1})^L$ of the dynamical system
%: the input sequence $\mathbf{x}(0)= (x_1(0),\ldots,x_L(0)) \in (\mathbb{S}^{d-1})^L$ is an initial condition which is evolved through the dynamical system:
\begin{align}\label{eq:mambaLN_dynamic}
    &\frac{d}{dt}x_{l}(t) = \mathbf{P}_{x_l(t)} \Big( \sum\limits_{j=1}^l A_{lj}(\mathbf{x}(t),t) x_{j}(t) \Big), \quad l=1,\ldots,L,\\
    &\mathbf{x}(0)=(x_1(0),\ldots,x_L(0)) \in (\mathbb{S}^{D-1})^L,
\end{align}
where
    \begin{itemize}
        \item for each $x \in \mathbb{S}^{D-1}$, the map $\mathbf{P}_{x} \colon \mathbb{R}^D \to T_{x}\mathbb{S}^{D-1}$ is the projection onto the tangent space at $x$ of the sphere $\mathbb{S}^{D-1}$ and it is defined by
        $\mathbf{P}_x(y) = y - \left< y,x\right>x$ for each $y \in \mathbb{R}^D$;

        \item the coefficient $A_{lj}$ is an $D \times D$ diagonal matrix $A_{lj} = \operatorname{diag}(A_{1,lj},\ldots,A_{D,lj})$ with
        \begin{align}\label{eq:mamba_interaction_coefficient}
            &A_{d,lj}(\mathbf{x}(t),t) =\left\{
            \begin{aligned}
                & x_l(t)^{\top} S_{BC}(t) x_l(t) \cdot \Delta_d(x_l(t),t), &\text{if } l = j,\\
                &x_l(t)^{\top} S_{BC}(t) x_j \cdot \Delta_d(x_j(t),t) \cdot \\
                & \quad \cdot \exp \Big( -a_d(t) \sum\limits_{k=j+1}^l \Delta_d(x_k(t),t) \Big), &\text{if } l > j;
            \end{aligned}
            \right.
        \end{align}

        \item the step size function at the $d$-th channel $\Delta_d \,:\, \mathbb{R}^D \times [0,+\infty) \to \mathbb{R}$ is defined by $\Delta_d(u,t) = \text{softplus}(S_{\Delta,d}(t) u)$,
        % \begin{align}
        %     \Delta_d(u,t) = \text{softplus}(S_{\Delta,d}(t) u) = \ln \left( 1+e^{S_{\Delta, d}(t) \cdot u} \right),
        % \end{align}
        for each $(u,t) \in \mathbb{R}^D \times [0,+\infty)$.
    \end{itemize}

\subsection{Mamba Interaction Coefficients Are More Complex Than Those in Causal Attention}

    In our continuous-time view for both Mamba and attention define dynamics, each $\frac{dx_l}{dt}$ is a linear combination of earlier tokens with interaction coefficients. For attention, the interaction coefficients are positive scalars as they are normalized by softmax.
    These positive scalar interaction coefficients makes the system comparatively structured and amenable to existing analysis (see \cite{karagodin2024clustering}). 
    
    In contrast, the interaction coefficients of Mamba are far more complicated than those in the causal attention setting.
    Indeed, the Mamba interaction coefficients are $A_{lj}(\mathbf{x}(t),t)$ which are given in equation~\eqref{eq:mamba_interaction_coefficient} are full matrices with no positivity or definiteness guarantees.
    As a consequence, almost all tools and conclusions from the attention setting, including hemisphere-based Lyapunov methods~\cite{abellaconsensus} and Shub-type genericity arguments~\cite{karagodin2024clustering}, break down.
\section{Theoretical Results for Mamba Dynamic}
\label{sec:theoretical-justifications}
%\vspace{-0.2cm}

Our goal is to study the dynamical properties of tokens in the internal representations of deep Mamba models.
In the continuous-time limit, this amounts to analyzing the solution trajectories of the system~\eqref{eq:mambaLN_dynamic} and characterizing the limiting set
$\lim_{t\to +\infty}\{x_1(t),\ldots,x_L(t)\}$.
In this section, we focus on a simplified Mamba setting in which the model parameters are assumed to be time-independent. 
An extension to the time-dependent parameter regime is discussed in Section~\ref{sec:time-dependence}.
% Extensions to the time-dependent parameter regime and to autoregressive sequence models are discussed in Sections~\ref{sec:time-dependence} and~\ref{sec:beyond}, respectively.
Intuitively, we will prove the following:

\begin{observation}\label{observation}
    \begin{itemize}
        \item[($i$)] The first token $x_1(t)$ has privileged dynamics: its trajectory determines a discrete set of possible limit points, primarily the coordinate unit vectors $\pm e_d$, which serve as attractors (Theorems~\ref{thm:first_token_D=2_main} and~\ref{thm:first_token_main}).
        Here, $e_d$ is the coordinate unit vector with $1$ at the $d$-th coordinate and $0$ elsewhere.
        
        \item[($ii$)] Subsequent tokens $x_l(t)$ with $2 \leq l \leq L$ are strongly influenced by the first token, often exponentially converging toward the same attractor, thereby creating clusters of tokens in the final layers (see Theorem~\ref{thm:all-conv_main} and Proposition~\ref{prop:positive} for the time-indepedent parameter setting and Theorem~\ref{thm:nonautonomous2_main} for a time-dependent analogue).
    \end{itemize}
\end{observation}

% \vspace{-0.2cm}
% \section{Theoretical Results for Mamba -- First Token's Influence}
% \label{sec:theoretical-justifications}
% \vspace{-0.2cm}

% In this subsection, we provide a theoretical justification for Observation~\ref{observation}. 
% In particular, we characterize the possible limit points of the first token in a pretrained deep Mamba model in Subsection~\ref{sec:first_token_main}, prove that all other tokens will exponentially converge to certain attractors corresponding to the limit points of the first token in Subsection~\ref{sec:other_tokens}, and extend our convergence results beyond Mamba, to general autoregressive sequence models in Subsection~\ref{sec:beyond}. 

The study of the convergence of all tokens based on the dynamical system~\eqref{eq:mambaLN_dynamic} without any additional assumption on the model parameters is challenging. 
To make the proofs feasible while still maintaining the generality practicality of our theoretical findings, we will first assume that all parameters $S_{BC}, S_{\Delta,d}$ and $a_d$ are time-independent in Sections~\ref{sec:first_token_main} and~\ref{sec:other_tokens}.
Next, a generalization into the case of time-dependent parameters will be presented in Section~\ref{sec:time-dependence}.
All proofs can be found in Appendix.
% {\color{red} ...need the case of time-dependent parameters...}

These theoretical observations will be empirically validate using practical Mamba architectures, considering both the continuous-time setting (in Subsection~\ref{sec:simulation}) and the discrete-time setting (in Subsection~\ref{sec:verification}).

%%%%%%%%%first tokens%%%%%%%%%%%%%

%\vspace{-0.2cm}
\subsection{Possible Limit Points of the First Token}\label{sec:first_token_main}
%\vspace{-0.2cm}

% {\color{red} In this subsection, all parameters $S_{BC}, S_{\Delta,d}$ and $a_d$ are assumed to be time-independent.
% The case when these parameters are time-dependent will be considered in Subsection~\ref{sec:collapse}.}

Let us consider the differential equation governing the trajectory of the first token, which is the first equation in the dynamical system~\eqref{eq:mambaLN_dynamic}, given by
\begin{align}\label{eq:first_token}
    &\frac{d}{dt} x_1(t) = \mathbf{P}_{x_1(t)}\big(A_{11}(\mathbf{x}(t)) x_1(t)\big),\\
    &x_1(0) \in \mathbb{S}^{D-1},\nonumber
\end{align}
where
\begin{align*}
    A_{11} &= \operatorname{diag}(A_{1,11}, A_{2,11}, \cdots, A_{D,11}), \text{ and}\\
    A_{d,11}(\mathbf{x}(t)) &= x_1(t)^{\top} S_{BC} x_1(t) \cdot \Delta_d(x_1(t)).
\end{align*}
% \[
% A_{11} = \operatorname{diag}(A_{1,11}, A_{2,11}, \cdots, A_{D,11}),\] 
% and
% \[A_{d,11}(\mathbf{x}(t)) = x_1(t)^{\top} S_{BC} x_1(t) \cdot \Delta_d(x_1(t)).
% \]
% and $A_{d,11} = \langle x_1, S_{BC} x_1 \rangle \Delta_d(x_1)$. 
A limit point must be an equilibrium point.
We will determine the set of equilibrium points of differential equation~\eqref{eq:first_token} in the following proposition.

\begin{proposition}[{All possible equilibrium points of the first token}]\label{prop:x1*_main}
   A vector $y \in \mathbb{S}^{D-1}$ is an equilibrium of equation~\eqref{eq:first_token} 
if and only if it belongs to one of the following types:
\begin{itemize}
   \item[] \textbf{Type 1.} 
   $y \in  
       \bigcup_{A \subset \{1,2,\ldots,D\}} \mathcal{H}_{\Delta,A}$, where for each subset $A \subset \{1,\ldots,D\}$, we denote by $\mathcal{H}_{\Delta,A} :=
          \left< 
            \{\, S_{\Delta,j} - S_{\Delta,\min A}, \, j \in A \,\} 
           \cup \{ e_k,\,k\in A\}\right>^{\perp}$. 
    %        the orthogonal complement of the vector space in $\mathbb{R}^D$ spanned by the differences
    % $\{\, S_{\Delta,j} - S_{\Delta,\min A} : j \in A \,\}$ and 
    % for some proper subset $A \subset \{1,\ldots,D\}$.
    % $y \in \mathcal{H}_{\Delta} := 
    %    \bigcup_{A \subset \{1,2,\ldots,D\}}
    %    \Bigl(
    %       \bigl\langle 
    %         \{\, S_{\Delta,j} - S_{\Delta,\min A} \;\big|\; j \in A \,\} 
    %       \bigr\rangle^{\perp}
    %       \cap
    %       \{\, y_k = 0 \;\; \text{if } k \notin A \,\}
    %    \Bigr).
    % $
    % which is the orthogonal complement of the span generated by 
    % $\{\, S_{\Delta,j} - S_{\Delta,\min A} : j \in A \,\}$ 
    % for some proper subset $A \subset \{1,\ldots,D\}$.
    % The vector $y$ may have some zero coordinates, but restricted to its nonzero entries we obtain
    % \[
    %     y^{(\ne 0)} = (y_{k_1},\dots, y_{k_\ell}) \in \mathbb{S}^{\ell-1},
    % \]
    % which lies in the common kernel of the column differences
    % \[
    %     S_{\Delta,k_j} - S_{\Delta,k_1}, \qquad j=1,\dots,\ell,
    % \]
    % for the active indices $\{k_1,\dots,k_\ell\}$.

    \item[] \textbf{Type 2.} $y \in \mathcal{H}_{S_{BC}}:= \{u \in \mathbb{R}^{D} \,|\, u^{\top} S_{BC} \, u = 0\}$, which is the null space of the quadratic surface defined by $S_{BC}$.
    % The vector $y$ satisfies
    % \[
    %     y^{\top} S_{BC}\, y = 0.
    % \]
\end{itemize}

If $S_{BC}$ is definite, then $\mathcal{H}_{S_{BC}} = \emptyset$ and Type~2 equilibria do not exist.  
If, furthermore, the columns of $S_{\Delta}$ are linearly independent, then the Type~1 equilibria collapse to exactly two antipodal points $y = \pm \xi$,
% \[
% y^{(\neq 0)} = \pm \xi,
% \]
where $\xi$ is the unique unit vector in the union $\bigcup_{A \subset \{1,2,\ldots,D\}} \mathcal{H}_{\Delta,A}$.
\end{proposition}

Not every equilibrium point is stable, and therefore not every equilibrium point can arise as a limit point.
Even for the first token, analyzing stability near equilibria in order to characterize the limit points of equation~\eqref{eq:first_token} is delicate and technically involved.
This difficulty already appears in the two-dimensional case $D=2$: depending on the initial data, the first token may converge to different limiting states, as we see in the following theorem.

\begin{theorem}[All possible limit points of the first token in two dimensions]
\label{thm:first_token_D=2_main}
    Assume that $D=2$ and let
    $y(t)=(y_1(t),y_2(t))
    \in C^{\infty}([0,+\infty),\mathbb{S}^1)$
    be a solution of differential equation~\eqref{eq:first_token}.
    Assume further that
    \begin{itemize}
      \item[($i$)]
      $S_{BC}=\operatorname{diag}(\lambda_1,\lambda_2)$
      with $\lambda_1\neq\lambda_2$ or
      $\lambda_1=\lambda_2\neq0$;
      \item[($ii$)]
      $S_{\Delta,2}\neq S_{\Delta,1}$.
    \end{itemize}
    In this case, the circle $\mathbb{S}^1$ intersects the line
    $\langle S_{\Delta,2}-S_{\Delta,1}\rangle^{\perp}$
    at two antipodal points, say $\pm\xi$.
    Then $\lim_{t\to+\infty}y(t)$ exists and belongs to the set
    \begin{align*}
        \Omega
        &=
        \{(0,\pm1),(\pm1,0)\}
        \cup \{\pm\xi\} \\
        &\quad\cup
        \left\{
            (\cos\varphi_\star,\sin\varphi_\star):
            \varphi_\star\in[0,2\pi),\ 
            \lambda_1\cos^2\varphi_\star
            +\lambda_2\sin^2\varphi_\star=0
        \right\}.
    \end{align*}
\end{theorem}

\begin{remark}[Possible limit points of the first token in practical setting]
    Theorem~\ref{thm:first_token_main} indicates that the first token converges to either of the coordinate unit vectors $\pm e_m$. 
    %consistent with our observation~\ref{observation}. 
    In contrast, Theorem~\ref{thm:first_token_D=2_main} reveals additional possible limit points, which are difficult to detect in simulations with randomly chosen model parameters.
    In practice, even when the assumptions of Theorems~\ref{thm:first_token_main} and~\ref{thm:first_token_D=2_main} do not strictly hold, we still observe that the first token typically converges to one of the coordinate unit vectors $\pm e_m$ (see Figure~\ref{fig:token_evolution} below).
\end{remark}

We leave the proof of the above proposition, as well as the study of the stability of these equilibrium points, to Appendix~\ref{sec:first_token}.
The following theorem proves that the first token $x_1(t)$ likely converges to a limit point of the form $\pm e_d$ for some $d=1,\ldots,D$.
% In Observation~\ref{observation}~($i$), we observe that the first token likely converge to a limit point of the form $\pm e_d$ for some $d=1,\ldots,D$. The following theorem provide a justification for this observation.

% \begin{theorem}[A justification for Observation~\ref{observation}~($i$) -- Convergence to $\pm e_d$ of the first token]\label{thm:first_token_main}
\begin{theorem}[First token likely converges to $\pm e_d$ in general]\label{thm:first_token_main}
    Let $y(t)=(y_1(t),\ldots,y_D(t)) \in C^{\infty}([0,+\infty),\mathbb{S}^{D-1})$ be a solution of differential equation~\eqref{eq:first_token}.
    Assume that the symmetric part 
    of $S_{BC}$ is positive definite, and $S_{\Delta,m} = \sign(y_m(0)) \cdot e_m$ for each $m$ such that $y_m(0) \neq 0$.
   Suppose that there exists $d_0\in\{1,..., D\}$ such that
    $$|y_{d_0}(0)|>|y_d(0)|>0,$$
    for every $d\neq d_0$.
    Then
    \[
        \lim_{t \to \infty} y(t) = \pm e_{d_0},
    \]
    where the sign is $\textnormal{sign}(y_{d_0}(0))$.
\end{theorem}

\begin{comment}
\begin{remark}
We provide theoretical support for the simulations in Figure~\ref{fig:token_evolution} by analyzing a canonical special case in which $S_{\triangle}$ is a one-hot matrix.
In high-dimensional settings, the columns of $S_{\triangle}$ are typically weakly correlated; in particular, pairwise cosine similarities are small with high probability due to concentration phenomena. In this first theoretical study, we neglect these small correlations and set them to zero in order to isolate the core dynamical mechanism. Importantly, the coordinates of the first token always evolve according to a replicator-type dynamics (see \eqref{eq:y_i}).
When the effective fitness is monotone in each coordinate, this dynamics selects the dominant components, leading to collapse onto antipodal states or low-dimensional cluster configurations.
More generally, the replicator structure persists even in the absence of component-wise monotonicity; however, the resulting long-time behavior can be substantially more intricate and may converge to other types of equilibria, which are in general non-isolated (Proposition~\ref{prop:x1*_main}) and depend on the underlying data distribution (Theorem~\ref{thm:first_token_D=2_main}). Understanding the interplay between replicator-type dynamics and high-dimensional structure remains an interesting and challenging direction for future work.
\end{remark}
\end{comment}

% \vspace{-0.2cm}
\subsection{All Other Tokens are Attracted by the First Token}\label{sec:other_tokens}
% \vspace{-0.2cm}

Different from the first token, the problem of characterizing the possible limits points of the other token is far more challenge.
In this subsection, to make the proofs feasible while still maintaining the generality and practicability of our findings, beside of the time-independent parameters assumption, we will assume further that:
\begin{align*}
    &S_{\Delta,1}(t) = \ldots = S_{\Delta,D}(t) =: \alpha \in \mathbb{R}^D, \text{ and}\\
    &a_1(t)=\ldots=a_D(t) =: a \in \mathbb{R},
\end{align*}
    % $$S_{\Delta,1}(t) = \ldots = S_{\Delta,D}(t) =: \alpha \in \mathbb{R}^D$$ 
    % and
    % $$a_1(t)=\ldots=a_D(t) =: a \in \mathbb{R},$$
for any time $t \in [0,+\infty)$.
Under this assumption, we can rewrite dynamical system~\eqref{eq:mambaLN_dynamic} as:
\begin{align}\label{eq:new-eq_main}
    &\frac{d}{dt} x_l(t)=\sum_{j=1}^{l-1}A_{lj}(\mathbf{x}(t)) \cdot \mathbf{P}_{x_l(t)}(x_j(t)), \quad l=1,\ldots,L,
    % &\mathbf{x}(0)=(x_1(0),\ldots,x_L(0)) \in (\mathbb{S}^{D-1})^L,\nonumber
\end{align}
where the coefficients $A_{lj}(\mathbf{x}(t))$ are defined as
\begin{align}\label{eq:A_ij}
    A_{lj}(\mathbf{x}(t))= x_l(t)^{\top} S_{BC}x_j(t) \cdot \triangle(x_l(t))e^{-a \sum_{k=j+1}^l \triangle(x_k(t))},
\end{align}
% \[
% A_{ij}(\mathbf{x}(t))= x_i(t)^{\top} S_{BC}x_j(t) \cdot a_{ij}(\mathbf{x}(t)),
% \]
with %\(a_{ij}(\mathbf{x}(t))=\triangle(x_i(t))e^{-a \sum_{k=j+1}^i \triangle(x_k(t))}\), and 
\(\triangle(x_l(t))=\ln(1+e^{\alpha \cdot x_l(t)})\).
% Throughout this section, we will assume that $x(t)=(x_1(t),x_2(t),\ldots,x_L(t)) \in C^{\infty}([0,+\infty);\mathbb{R}^D)^L$ is a solution of the dynamical system~\eqref{new-eq}.

% In the remaining part of this section, we assume that $\mathbf{x}(t)=(x_1(t),x_2(t),\ldots,x_L(t)) \in C^{\infty}([0,+\infty);\mathbb{R}^D)^L$ is a solution of the dynamical system~\eqref{eq:new-eq_main}.
In this case, for the first token $x_1(t)$, we see that 
\[
      \frac{d}{dt} x_1(t)=A_{11}(\mathbf{x}(t),t) \cdot \mathbf{P}_{x_1(t)}(x_1(t))=0.
\]
This implies that $x_1(t)=x_1(0)$ is a constant function on $[0,+\infty)$.
We will present all possible limit points of $x_i(t)$ for $i= 2, \ldots, L$ in Theorem~\ref{thm:all-conv_main} and Proposition~\ref{prop:positive} below. % under more restrictive conditions.
% Furthermore, in Theorem~\ref{sec:collapse_appendix}, we provide a sufficient condition on the model parameters under which $\lim_{t \to +\infty}x_l(t)=x_1$ for all $l$.

% \red{To do: how to state the theorem in a nice way,suitable to put at the beginning for audiences?.. We also want to select which theorem to put from the theory appendix. also the last theorem we can maybe compare with the result in the MIT  group causal attention paper, reprove and get exp decay?}

% \subsection{More than one clustering case}

% \begin{theorem}[A justification for Observation~\ref{observation}~($ii$) -- First token exponentially attracts other tokens]\label{thm:all-conv_main}
\begin{theorem}[All tokens are exponentially attracted by the first token]\label{thm:all-conv_main}
Let $\mathbf{x}(t)= (x_1(t),\ldots,x_L(t)) \in C^{\infty}([0,+\infty), (\mathbb{S}^{D-1})^L)$ be a solution of dynamical system~\eqref{eq:new-eq_main}. Assume that $S_{BC}$ is symmetric and $\langle x_1(0),S_{BC}x_1(0)\rangle>0$.
Then, there exists a zero-measurable subset $\Omega$ of $(\mathbb{S}^{D-1})^{L-1}$  such that: if the initial point $(x_2(0),...,x_L(0))$ does not belong in $\Omega$  then all tokens converge exponentially to $\pm x_1(0)$, i.e.,
\[
\sum_{i=1}^L\lw|x_i(t)-s_i x_1(0)\rw|\lesssim e^{-C_0t},
\]
for some constant $C_0>0$ and $s_i \in \{+1, -1\}$ for each $i$.
\end{theorem}

\paragraph{Proof strategy.}
The proof is organized into four steps.

\emph{Step 1 (two-token reduction).}
We first analyze the simplest nontrivial interaction between the first two tokens $x_1$ and $x_2$. 
The key scalar is
$\bigl\langle x_2(t),\,S_{BC}x_1\bigr\rangle$,
whose sign controls the long-time behavior. A Grönwall-type estimate yields exponential convergence of $x_2(t)$ to $x_1(0)$.

\emph{Step 2 (global positivity regime).}
We then extend the two-token analysis to the full system under the global positivity assumption that
$\bigl\langle x_i(0),\,S_{BC}x_j(0)\bigr\rangle>0$ for all $i,j$.
The main tool is a box-invariance argument for the interaction variables
$\bigl\langle x_i(t),\,S_{BC}x_j(t)\bigr\rangle$,
showing that trajectories starting in the positive box $[0,1]^{L\times L}$ cannot exit it (by a contradiction argument based on uniqueness of ODE solutions). This yields exponential convergence of all downstream tokens to $\pm x_1(0)$.

\emph{Step 3 (removing positivity by induction).}
Next, we drop the positivity assumption and proceed inductively. Assuming that $x_1,\dots,x_{k-1}$ converge exponentially to $x_1$, we show that the interaction coefficient
$\bigl\langle x_k(t),\,S_{BC}x_1\bigr\rangle$
decays exponentially, implying that $x_k(t)$ converges to an equilibrium orthogonal to $S_{BC}x_1$.

\emph{Step 4 (nongenericity of orthogonal equilibria).}
Finally, we prove that the orthogonal-drift scenario from Step~3 is nongeneric. For a candidate equilibrium
$x_*=(x_1,\ldots,x_1,\xi)$, with $\xi\perp S_{BC}x_1$,
we linearize the flow at $x_*$ and show that the Jacobian $J(x_*)$ admits a positive eigenvalue
$\lambda_+=a_{L1}(x_*)>0$,
so that $x_*$ is a saddle point. The stable manifold theorem then implies that the set of initial conditions leading to the convergence of tokens to $x_*$ lies in a lower-dimensional manifold of Lebesgue measure zero.
This provides a complete proof of Theorem~\ref{thm:all-conv_main}.

Under slightly stronger assumptions, we can apply Theorem~\ref{thm:all-conv_main} to recover the single-cluster case stated below.

\begin{proposition}[Exponential alignment under positive angle conditions]\label{prop:positive}
Suppose the assumptions of Theorem~\ref{thm:all-conv_main} hold. 
In addition, assume that the initial configuration satisfies the positive angle condition:
\[
\langle x_i(0), S_{BC} x_j(0) \rangle > 0, \quad \forall\, i,j \in [1,L].
\]
Then all tokens $x_l(t)$ converge exponentially to the aligned state $x_1(0)$, that is,
\[
\sum_{i=1}^L \|x_i(t) - x_1(0)\| \;\lesssim\; e^{-C_0 t},
\]
for some constant $C_0 > 0$. Moreover, the positivity of interactions is preserved for all time:
\[
\langle x_i(t), S_{BC} x_j(t) \rangle > 0, \quad \forall\, t>0.
\]
\end{proposition}

\begin{remark}[{All token are exponentially attracted by the first token in practical settings}]
    Theorem~\ref{thm:all-conv_main} states that, under certain assumptions, each token converges exponentially either to $x_1(0)$ or to $-x_1(0)$, depending on the sign of the corresponding coefficient $s_i$. Thus, in this setting there are two possible clusters: one around $x_1(0)$ and one around $-x_1(0)$.
    Proposition~\ref{prop:positive} imposes stricter assumptions under which all tokens converge exponentially to $x_1(0)$. In this case, there is only one cluster.

    In practice, when the assumptions of Theorem~\ref{thm:all-conv_main} and Proposition~\ref{prop:positive} do not fully hold, it is possible to observe more than two clusters at the end of the dynamics (see for examples in Figure~\ref{fig:token_evolution} below). However, this occurs only rarely. Almost all trajectories still fall into one of the two dominant clusters predicted by Theorem~\ref{thm:all-conv_main}.
    
    % Theorem~\ref{thm:all-conv_main} together with Proposition~\ref{prop:positive} implies that all tokens converge exponentially to the limit points of the first token. 
    % %thereby confirming Observation~\ref{observation}~(ii). 
    % This, in turn, indicates that tokens in a deep Mamba model tend to form a small number of clusters in the final layers.
\end{remark}

%\vspace{-0.2cm}
\subsection{A generalization to time-dependent regime}\label{sec:time-dependence}

In this section, we discuss a generalization of Theorem~\ref{thm:all-conv_main} to the setting of time-dependent parameters.
% , focusing on the possible limit points of the token dynamics. 
Specifically, we allow the parameters to be nonconstant functions of time, subject to additional mild conditions ensuring that the proof strategy from the previous section remains valid.
% Throughout, we assume that $S_{BC}(t), \alpha(t)$ and $a(t)$ converge exponentially to $S_{\infty}$, $\alpha_{\infty}$ and $a_{\infty}$, respectively.

% In this section, we consider the problem \eqref{eq:new-eq_main} in the case where $S_{BC}, \alpha$ and $a$ are time-dependent. Moreover, 

We borrow the notation from the previous subsection.
To keep the proofs tractable while maintaining the generality and practical relevance of our results, we assume that the functions $S_{BC}(t)$, $\alpha(t)$, and $a(t)$ converge exponentially to $S_{\infty}$, $\alpha_{\infty}$, and $a_{\infty}$, respectively, when $t$ tends to infinity.
(In the previous section, these parameters are constant.)
In this case, the interaction coefficients are the matrices
\begin{align*}
    A_{lj}(t, x)= x_l^{\top} S_{BC}(t)x_j \cdot \triangle_t(x_l)e^{-a(t) \sum_{k=j+1}^l \triangle_t(x_k)},
\end{align*}
where \(\triangle_t(x_l)=\ln(1+e^{\alpha(t) \cdot x_l})\) for each $1\leq j<l\leq L$ and $x=(x_1,...,x_L)\in (\mathbb{S}^{D-1})^L$.
Under the above setting, we can generalize Theorem~\ref{thm:all-conv_main} to the time-dependent parameter regime as follows.

\begin{theorem}\label{thm:nonautonomous2_main} 
    Let $\mathbf{x}(t)= (x_1(t),\ldots,x_L(t))$ be a solution to the dynamical system
    \begin{equation}
        \begin{cases}\label{eq: nonautonomous2}
        \frac{d}{dt} x_l(t)=\sum_{j=1}^{l-1}A_{lj}(t+c, \mathbf{x}(t)) \cdot \mathbf{P}_{x_l(t)}(x_j(t)),\\
        \mathbf{x}(0)=(x_1(0),\ldots,x_L(0)) \in (\mathbb{S}^{D-1})^L,
        \end{cases}
    \end{equation}
    for $l=1,\ldots,L$, and $c \in \mathbb{R}$ is an arbitrary constant.
    Assume that $S_{BC}(t)$ is symmetric for every $t$ and $S_{\infty}$ is positive definite.
    Then, there exists a zero-measurable subset $N_0$ of $(0, \infty)$
    satisfying the following property: if $c\in \mathbb{R}\setminus N_0$ 
    % and $x_1(0)\in \mathbb{S}^{D-1}$ 
    then
    there exists a  zero-measurable subset
    $\Omega$ of $(\mathbb{S}^{D-1})^{L-1}$ such that if the initial point $(x_2(0),...,x_L(0))$ does not belong in $\Omega$  then all tokens converge exponentially to $\pm x_1(0)$.
\end{theorem}

Theorem~\ref{thm:nonautonomous2_main} indicates that the downstream tokens $x_l(t)$, for $2\le l\le L$, are strongly driven by the initial direction of the first token $x_1(0)$. 
In particular, for generic initializations, these tokens converge exponentially to the same attractor $\pm x_1(0)$, which leads to the emergence of token clusters in the final layers. 
This conclusion corroborates the phenomenon described at the beginning of Section~\ref{sec:theoretical-justifications}. 
Further discussion and intuition for this result are provided in Appendix~\ref{sec:time-dependence_appendix}.

\begin{figure}
    \centering
    \begin{subfigure}[t]{0.32\linewidth}
        \centering
        \includegraphics[width=1\linewidth]{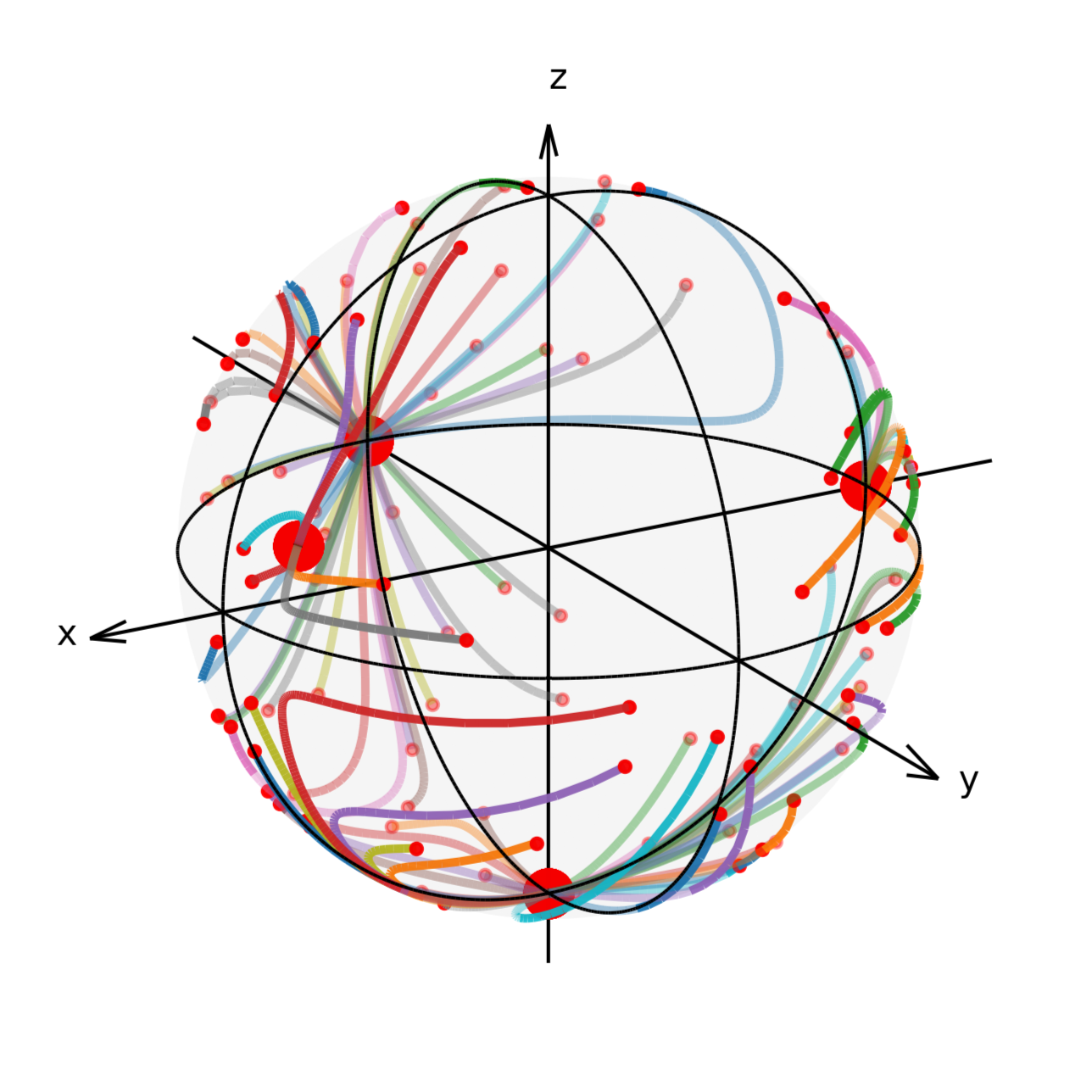}
        \caption{First-token trajectories (100 random initializations; time-independent  $a_d$, $S_{BC}$, $S_\Delta$).  94 trajectories converge to some coordinate unit vector $\pm e_d$.}
        \label{fig:firsttoken}
    \end{subfigure}\hfill
    \begin{subfigure}[t]{0.32\linewidth}
        \centering
        \includegraphics[width=1\linewidth]{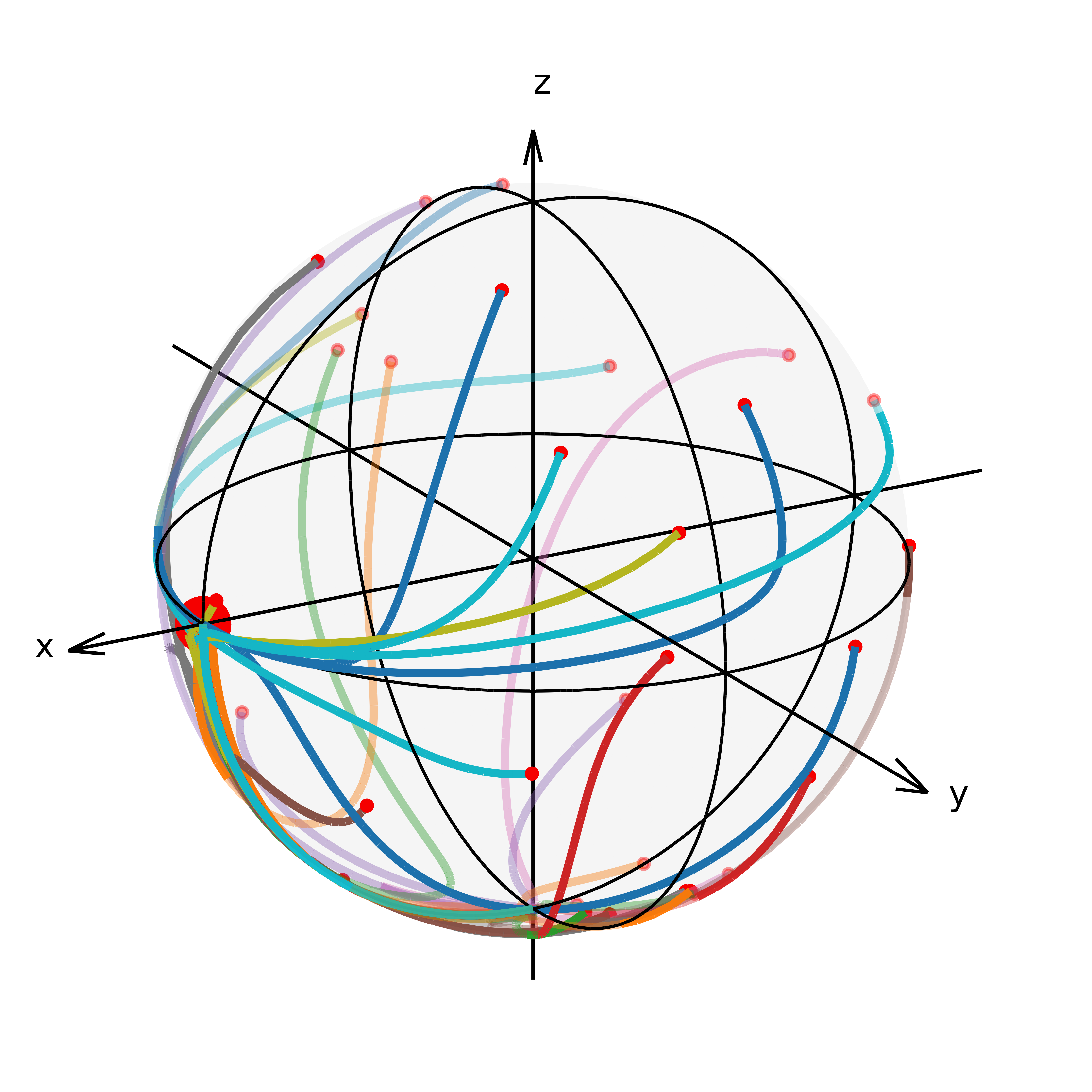}
        \caption{A single sequence of 32 tokens (time-independent $a_d$, $S_{BC}$, $S_\Delta$). Tokens converge toward a point at one coordinate unit vector $\pm e_d$ for some $d$.}
        \label{fig:allrandn}
    \end{subfigure}\hfill
    \begin{subfigure}[t]{0.32\linewidth}
        \centering
        \includegraphics[width=1\linewidth]{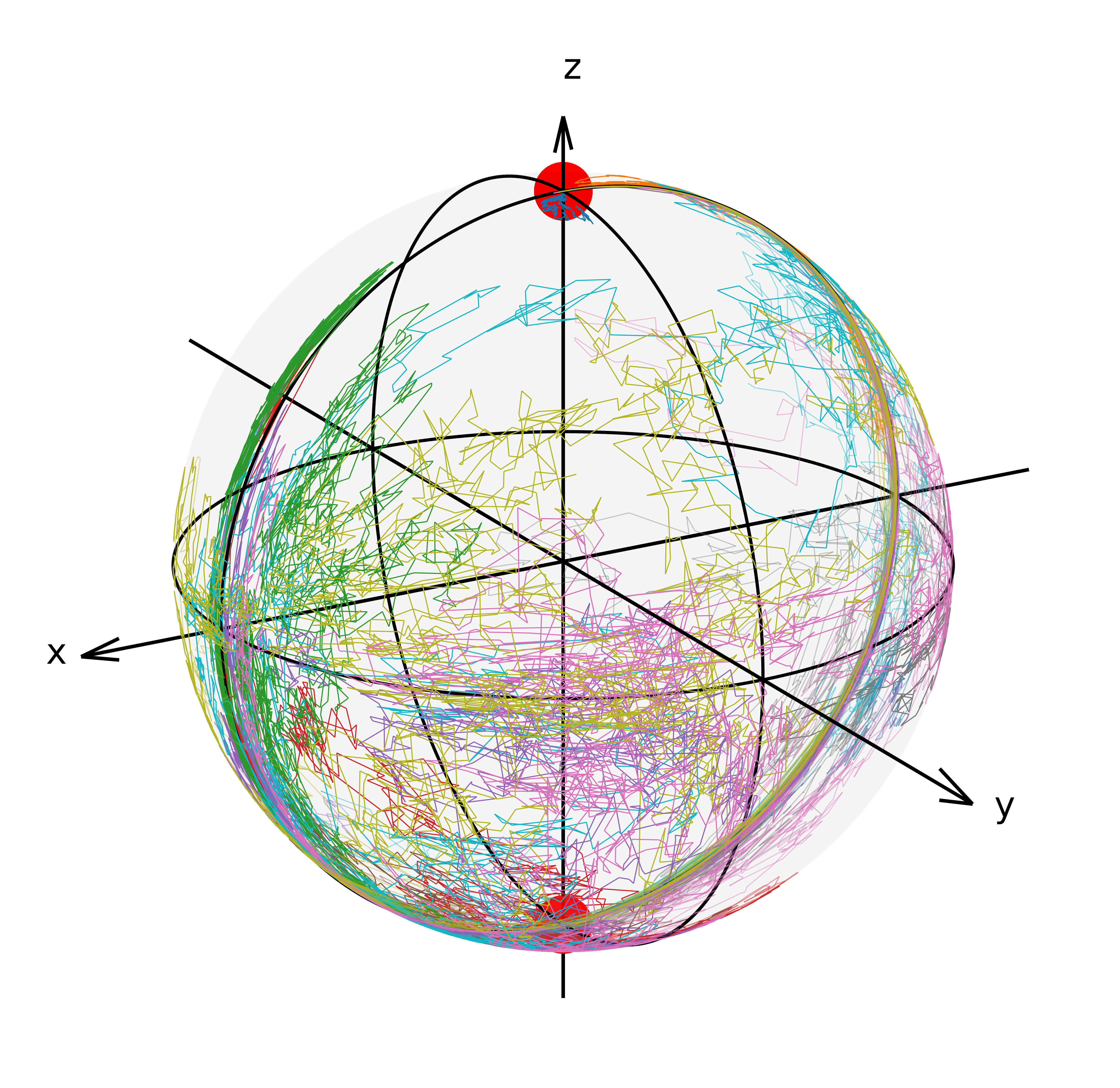}
        \caption{A single sequence of 16 tokens (time-dependent $a_d$, $S_{BC}$, $S_\Delta$). Tokens converge to points at some coordinate unit vectors $\pm e_d$ for some $d=1,\ldots,D$.}
        \label{fig:timedep}
    \end{subfigure}
    \caption{Simulated trajectories of dynamical system~\eqref{eq:mambaLN_dynamic} with random parameters. The initial points (denoted by small red points in (a) and (b)) are sampled via a uniform distribution on the sphere.
    In these settings, the first token converges to discrete attractors at the big red points (mostly coordinate unit vectors $\pm e_d$) and draws subsequent tokens toward the same points, producing clusters and illustrating the attention sink effect.
    }
    \label{fig:token_evolution}
    \vspace{-1em}
\end{figure}

%\vspace{-0.4cm}
\section{Alignment with Observed Behavior in Practical Mamba Settings}\label{sec:simulation}
%\vspace{-0.2cm}

We empirically validate the theoretical results established in the previous section using practical Mamba architectures, considering both the continuous-time setting (Subsection~\ref{sec:simulation}) and the discrete-time setting (Subsection~\ref{sec:verification}).
We study models with both time-dependent and time-independent parameters, and further investigate how practical architectural components of Mamba (such as pre-layer normalization and linear layers) impact the conclusions of our theory.

%\vspace{-0.3cm}
\subsection{Simulations}\label{sec:simulation}
%\vspace{-0.2cm}

% {\color{red} .. to be edited later ...}

% Before providing a rigorous theoretical analysis of the token dynamics of Mamba via dynamical system~\eqref{eq:mambaLN_dynamic}, 

To validate our theoretical results, we conduct extensive simulations of the dynamical system~\eqref{eq:mambaLN_dynamic} under random choices of model parameters, demonstrating that the empirical behaviors observed in the simulations are consistent with our theoretical predictions.
In particular, we numerically integrate the system to observe: (i) trajectories of first tokens across many random initializations with time-independent parameters $a_d$, $S_{BC}$, and $S_\Delta$; (ii) trajectories of all tokens in a single sequence with time-independent parameters $a_d$, $S_{BC}$, and $S_\Delta$; and (iii) trajectories of all tokens in a single sequence with time-dependent parameters $a_d$, $S_{BC}$, and $S_\Delta$.
% Figure~\ref{fig:token_evolution} demonstrates the expected results of the trajectories of the tokens under these settings.

Figure~\ref{fig:token_evolution}, which demonstrates the expected results of the trajectories of the tokens under the above settings, reveals two robust patterns. 
First, the initial token exhibits privileged dynamics and converges to a discrete set of attractors, namely the coordinate vectors $\pm e_d$, in agreement with Theorems~\ref{thm:first_token_main} and~\ref{thm:first_token_D=2_main}. Second, the downstream tokens are strongly driven by this first-token trajectory and typically converge to the same attractor, producing token clusters in the final layers, consistent with Theorem~\ref{thm:all-conv_main}, Proposition~\ref{prop:positive}, and Theorem~\ref{thm:nonautonomous2_main}.
These robust patterns strongly support Observation~\ref{observation}.

% \begin{observation}\label{observation}
%     From Figure~\ref{fig:token_evolution}, two consistent patterns can be observed:
%     \begin{itemize}
%         \item[($i$)] The first token has privileged dynamics: its trajectory determines a discrete set of possible limit points, primarily the coordinate unit vectors $\pm e_d$, which serve as attractors.
%         Here, $e_d$ is the coordinate unit vector with $1$ at the $d$-th coordinate and $0$ elsewhere.
%         This observation confirms our findings in Theorems~\ref{thm:first_token_main} and~\ref{thm:first_token_D=2_main}.
        
%         \item[($ii$)] Subsequent tokens are strongly influenced by the first token, often converging toward the same attractor, thereby creating clusters of tokens in the final layers.
%         This observation confirms our findings in Theorem~\ref{thm:all-conv_main}, Proposition~\ref{prop:positive}, and Theorem~\ref{thm:nonautonomous2_main}.
%     \end{itemize}
% \end{observation}

% The above observation directly supports the claim that the first token acts as a global attractor and explains phenomena like the attention sink effect in Mamba/causal models (see Section~\ref{sec:experiment} for further details).
% The next section provides a theoretical explanation for the above observation.

%\vspace{-0.3cm}
\subsection{Empirical Validation with Full Practical Setting}\label{sec:verification}
%\vspace{-0.2cm}

To validate our theoretical results on practical models, we conduct experiments on simplified versions of the Mamba-2 block. Specifically, we employ the configuration with shared-weight blocks. Within each block, we omit the causal convolution, input projection, and output projection.

\begin{figure}
    \centering  %\vspace{-0.5cm}
    \includegraphics[width=0.6\linewidth]{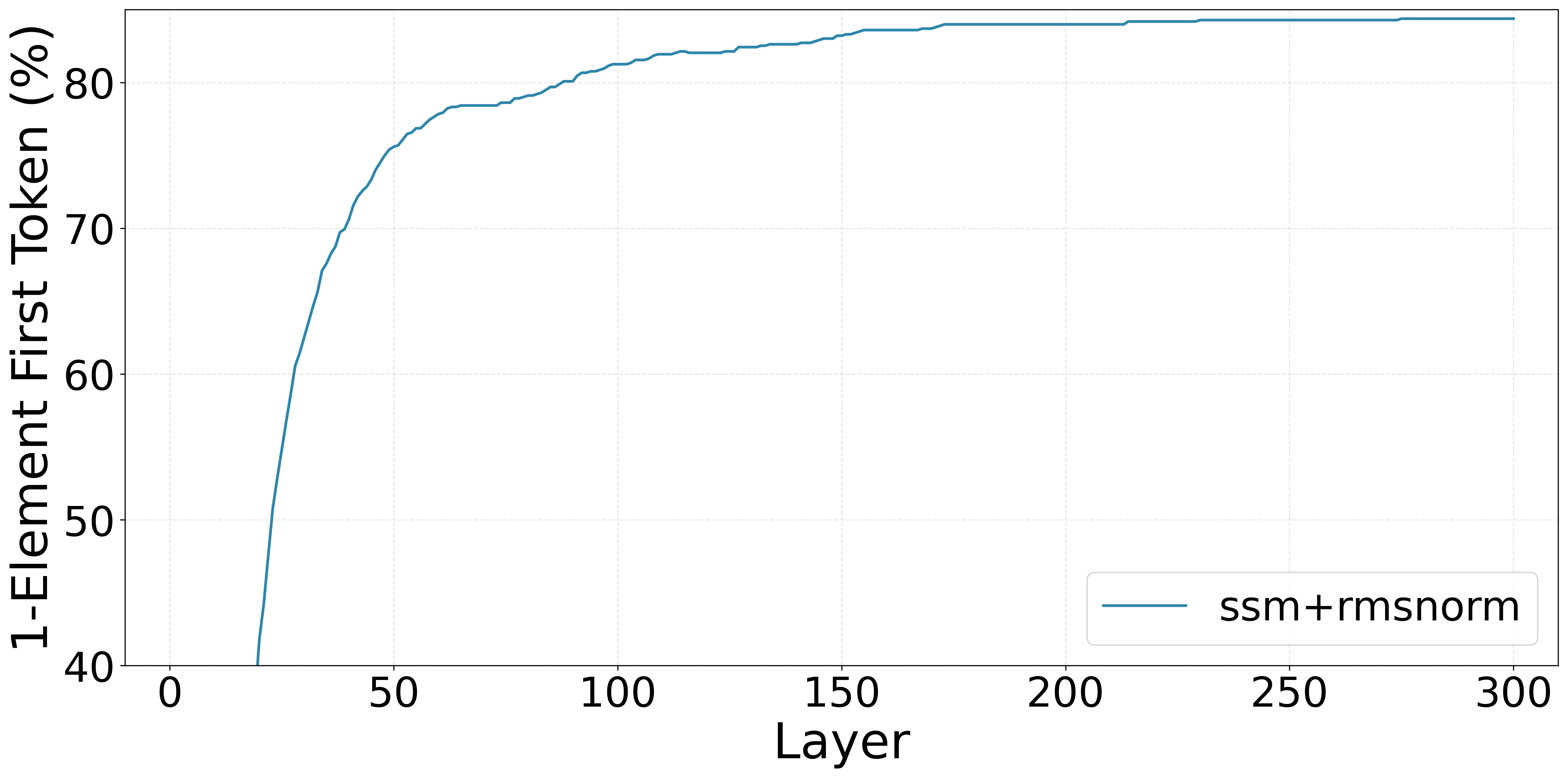}
    \caption{Percentage of first tokens exhibiting a single non-zero element (threshold $1\times 10^{-4}$) across simplified Mamba-2 blocks with shared weights, after omitting causal convolution, input/output projections, activations, and inner residuals. Most tokens converged to 1-element tokens, consistent with simulations.} \label{fig:sparse_token_percentage}
    % \vspace{-0.1cm}
\end{figure}

We process a batch of 64 sequences through the model. We compute the percentage of sequences where the first token has only one non-near-zero element (threshold $1\times10^{-4}$). The results in Figure~\ref{fig:sparse_token_percentage} reveal that most first tokens converge to a representation with only one non-zero element. This outcome aligns with our theoretical predictions and the simulation in Figure~\ref{fig:firsttoken}.

\begin{figure}
    \centering %\vspace{-0.5cm}
    \begin{subfigure}[c]{0.22\textwidth}
        \includegraphics[width=\textwidth]{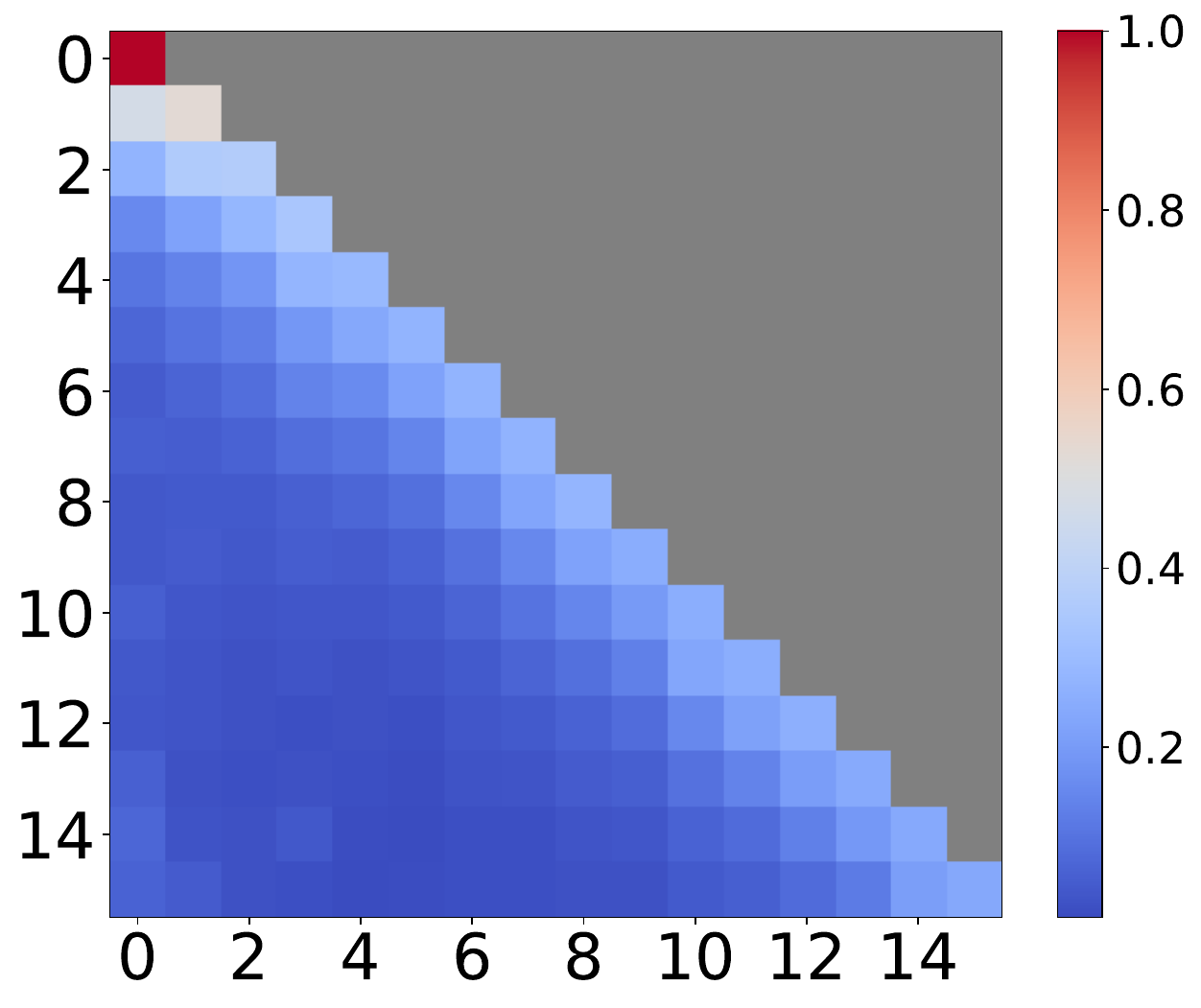}
        \caption{Layer 1, Channel 8}
        \label{fig:attn-layer1}
    \end{subfigure}
    \hfill
    \begin{subfigure}[c]{0.22\textwidth}
        \includegraphics[width=\textwidth]{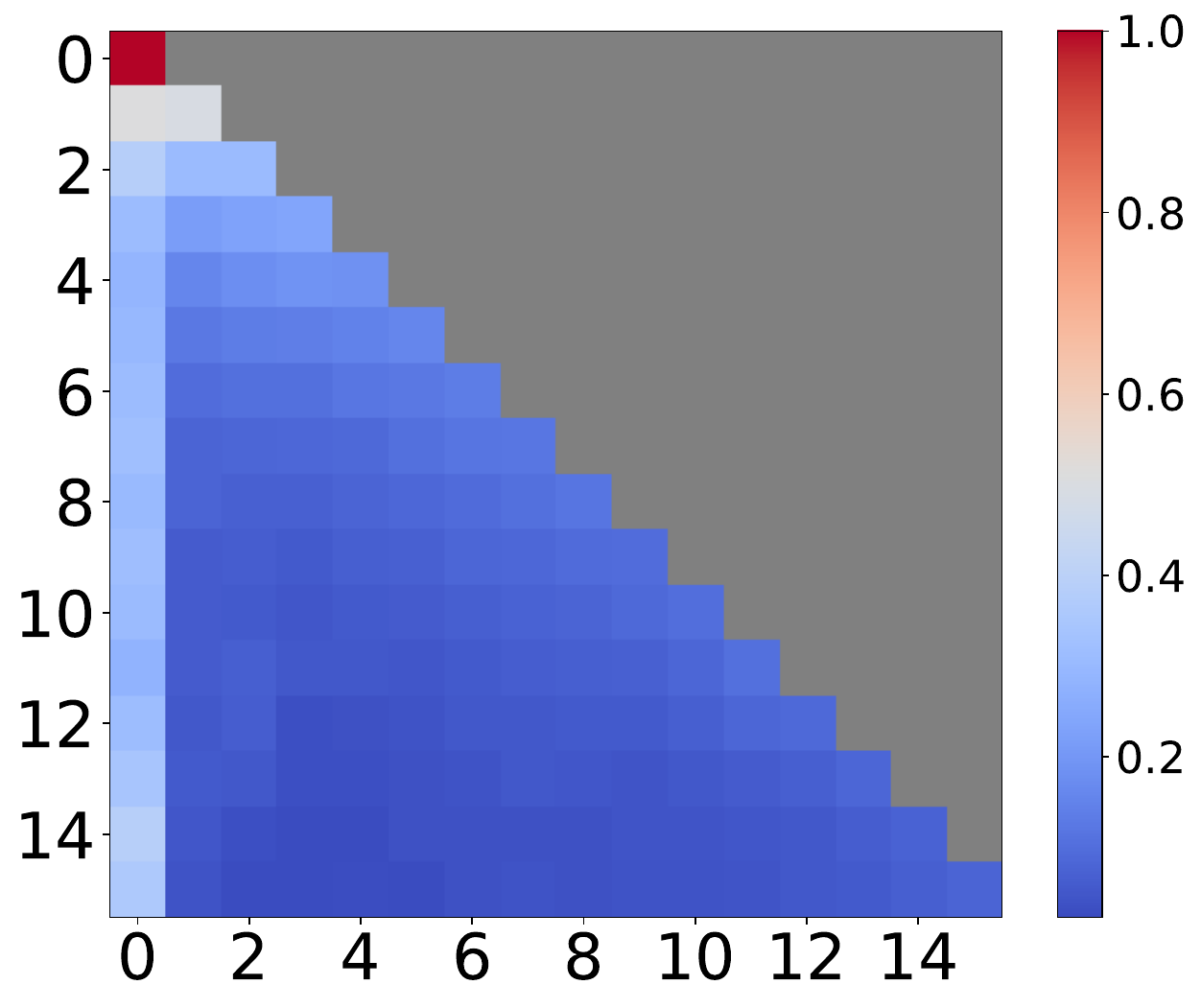}
        \caption{Layer 15, Channel 8}
        \label{fig:attn-layer15}
    \end{subfigure}
    \hfill
    \begin{subfigure}[c]{0.22\textwidth}
        \includegraphics[width=\textwidth]{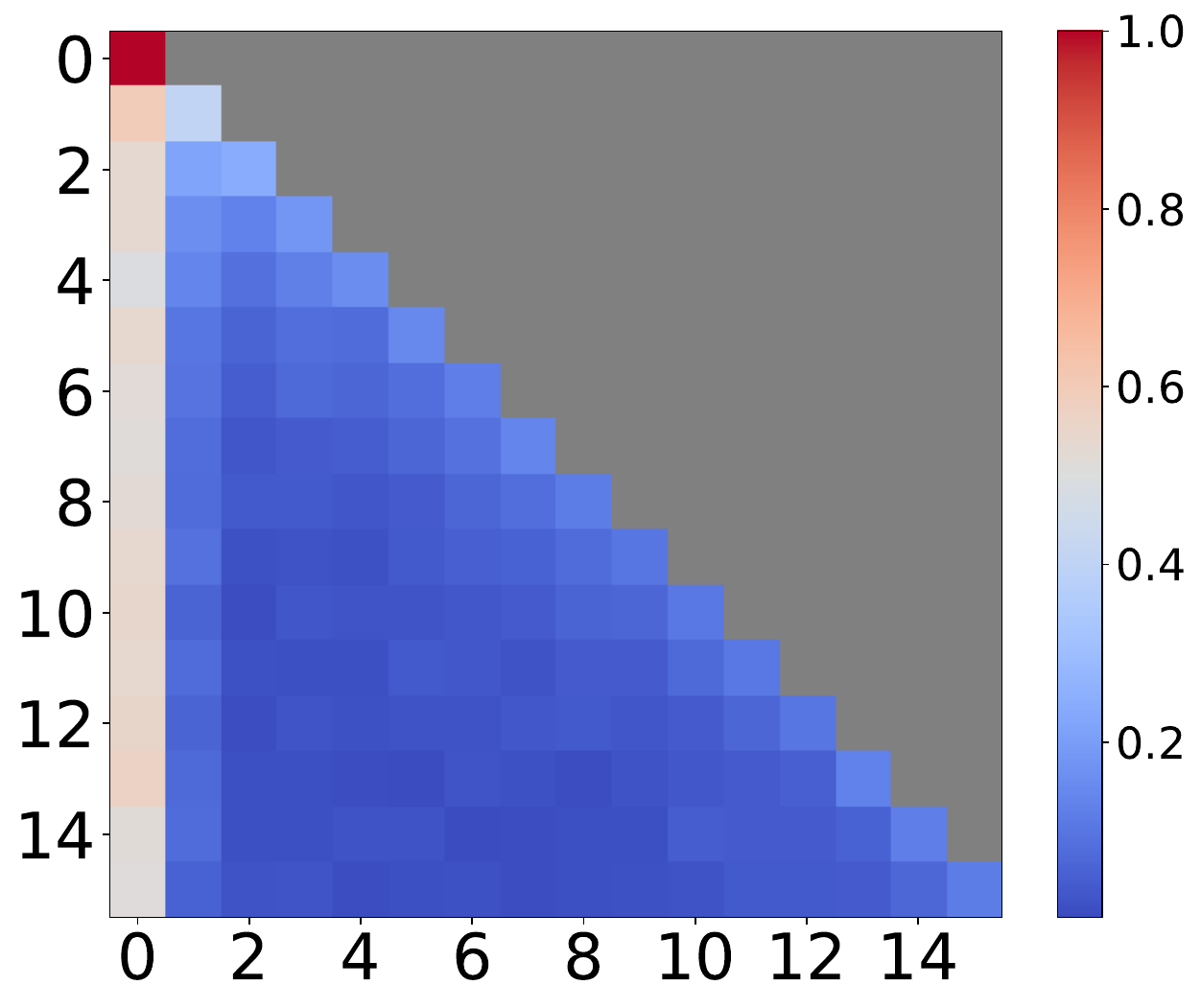}
        \caption{Layer 46, Channel 8}
        \label{fig:attn-layer46}
    \end{subfigure}
    \hfill
    \begin{subfigure}[c]{0.22\textwidth}
        \includegraphics[width=\textwidth]{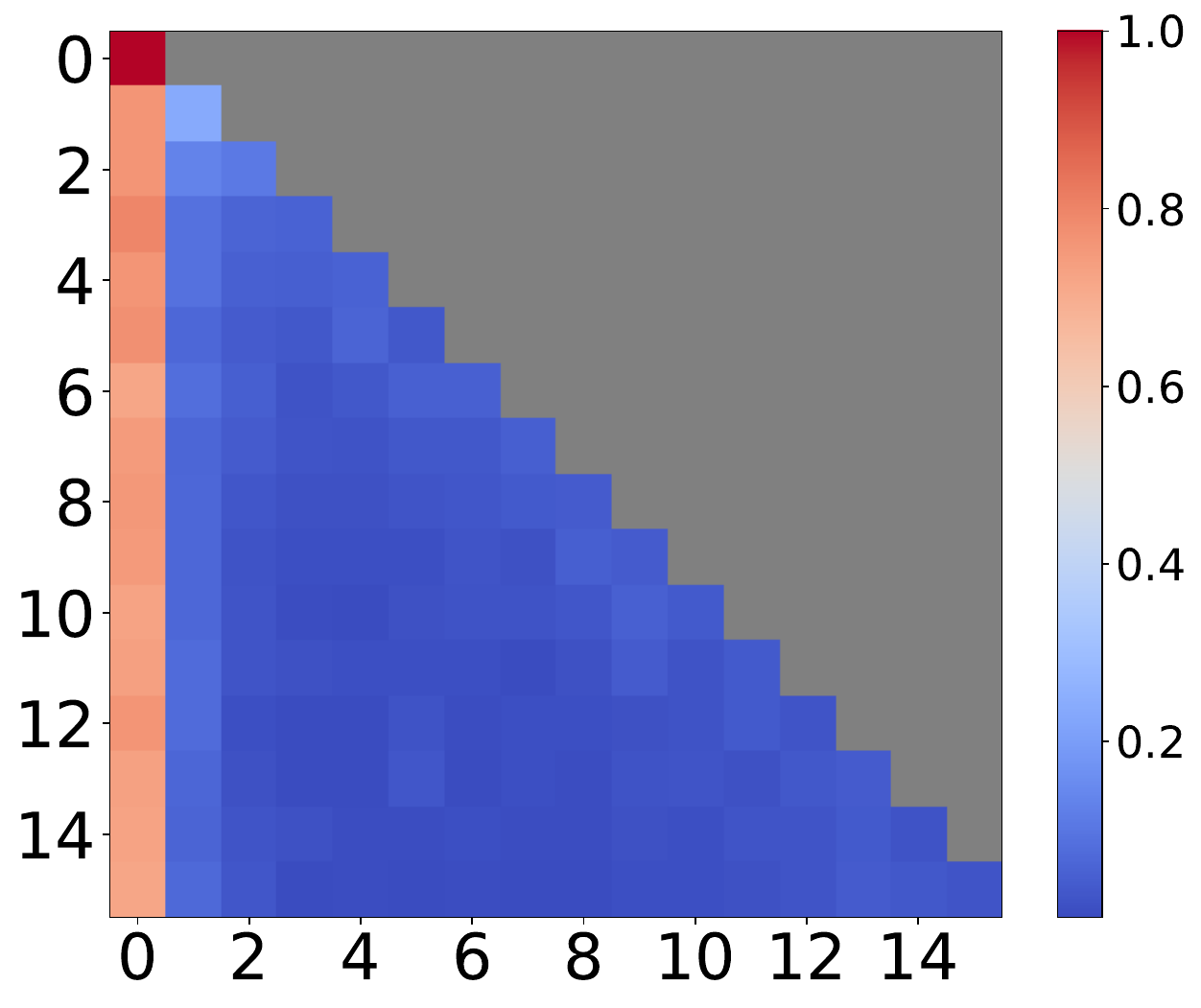}
        \caption{Layer 60, Channel 8}
        \label{fig:attn-layer60}
    \end{subfigure}
    \caption{Attention concentration on the first token across different depths of the Falcon-Mamba-7B model \citep{zuo2024falcon}. Each subfigure shows the hidden attention matrix for a specific layer: (a) Layer 1, (b) Layer 15, (c) Layer 46, and (d) Layer 60. In early layers, tokens distribute attention more uniformly, but as depth increases, nearly all tokens concentrate disproportionately on the first token. This illustrates the emergence of an attention sink in state-space models.} %, analogous to the phenomenon previously observed in transformer models.}
    \label{fig:attn-sink}
    \vspace{-1em}
\end{figure}

%\vspace{-0.4cm}
\section{An application -- Predict Attention Sink}\label{sec:experiment}
%\vspace{-0.2cm}

Observation~\ref{observation}, together with the theoretical justification provided in the previous section, suggests that the first token should be regarded with particular importance, as it serves as an attractor for subsequent tokens.  
This insight provides a principled explanation of the attention sink effect in Mamba models through the attracting role of the first token.

% Building on this insight, we provide in Subsection~\ref{sec:attention-sink} a principled explanation of the attention sink effect in Mamba models through the attracting role of the first token.  
% Moreover, in Subsection~\ref{sec:random-reordering-token}, we propose a practical strategy for Mamba models: randomly reordering tokens during training to diversify clustering around a small set of limit points, thereby enhancing model performance. We provide the details of experiment setup and evaluation benchmark of both experiments in Appendix~\ref{sec:experiment_appendix}.

% we empirically validate this claim by showing that the attention sink phenomenon \citep{xiao2024efficient, gu2025attention}, originally identified in transformer language models, is also observed in the hidden attention matrix of state-space language models as defined in equation~\ref{eq:hidden-attn}, for which we provide a complete theoretical explanation. Furthermore, we propose a reordering strategy for the Mamba vision model that diversifies the first token, resulting in consistent improvements in model performance.

% \begin{enumerate}
%     \item Tung: simulation
%     \item Khoi: predict attention sink and explain it
%     \item Tung: random reordering tokens
% \end{enumerate}
% }

% \subsection{Predict Attention Sink}
% \label{sec:attention-sink}

We analyze the hidden attention matrix in equation~\eqref{eq:convolution_rule_D} for the Falcon-Mamba-7B model \citep{zuo2024falcon} to quantify how tokens attend to one another. The results in Figure~\ref{fig:attn-sink} reveal that the concentration of attention on the first token consistently increases with depth, and in the final layers it dominates nearly all other tokens. This empirical trend validates our theoretical justification for token dynamics and, importantly, represents the first observation of the attention sink phenomenon in state-space models. While attention sink was originally identified in transformer-based architectures \citep{xiao2024efficient}, we have showed that an analogous effect emerges in Mamba models as predicted from Observation~\ref{observation}. %, and we explain this behavior within our token dynamic framework. 
%To complete the picture, we provide a theoretical justification for the emergence of attention sink in SSMs, \textcolor{red}{Thieu: add theoretical justification for attention sink here}.

\begin{figure}
    \centering
    % \vspace{-0.2cm}
    \begin{subfigure}[c]{0.5\textwidth}
        \centering
        \includegraphics[width=0.9\textwidth]{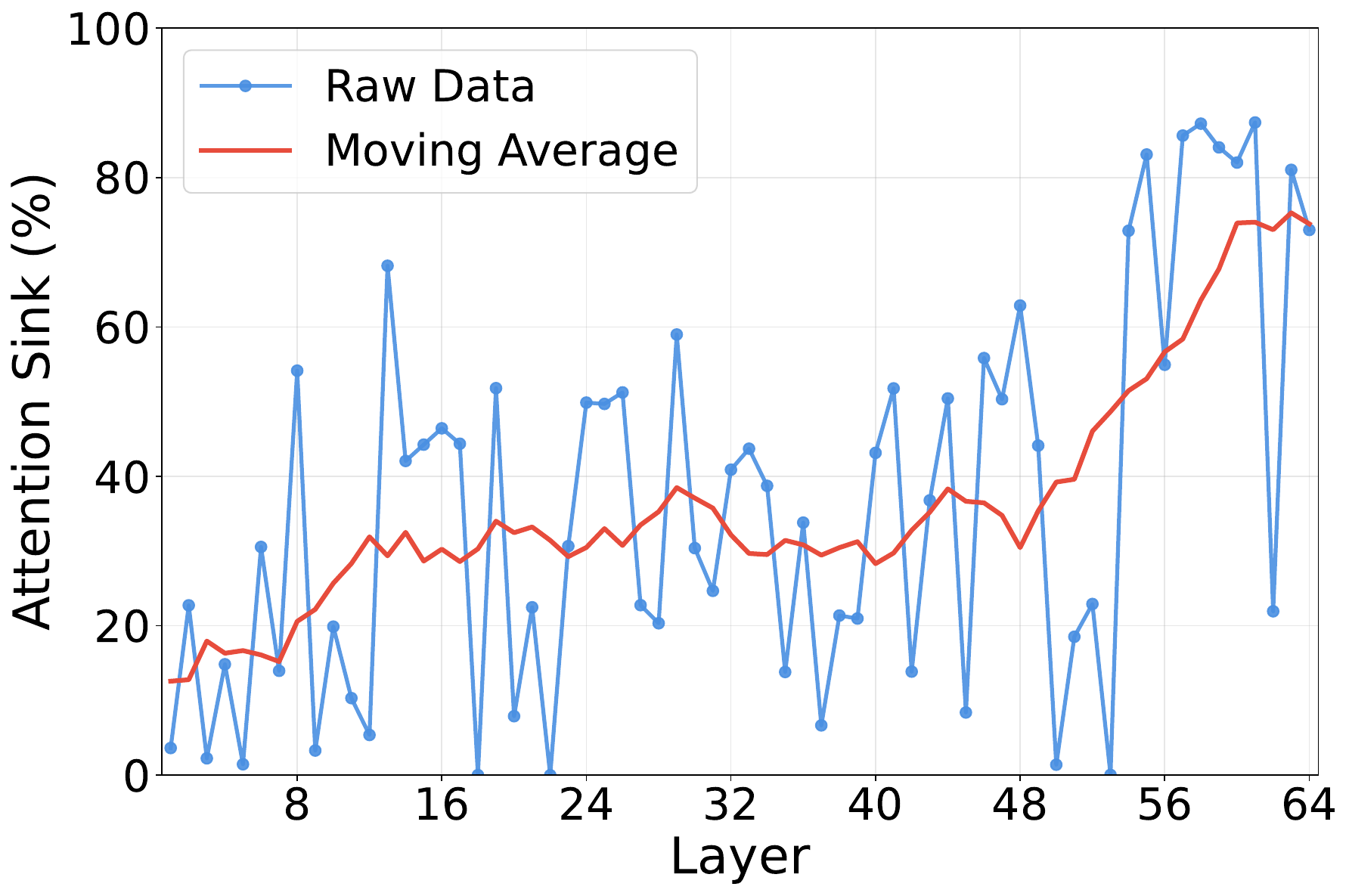}
        \caption{Attention sink (\%) across layers for the first token.}
        \label{fig:attn-sink-token0}
    \end{subfigure}
    \hfill
    \begin{subfigure}[c]{0.45\textwidth}
        \centering
        \includegraphics[width=1\textwidth]{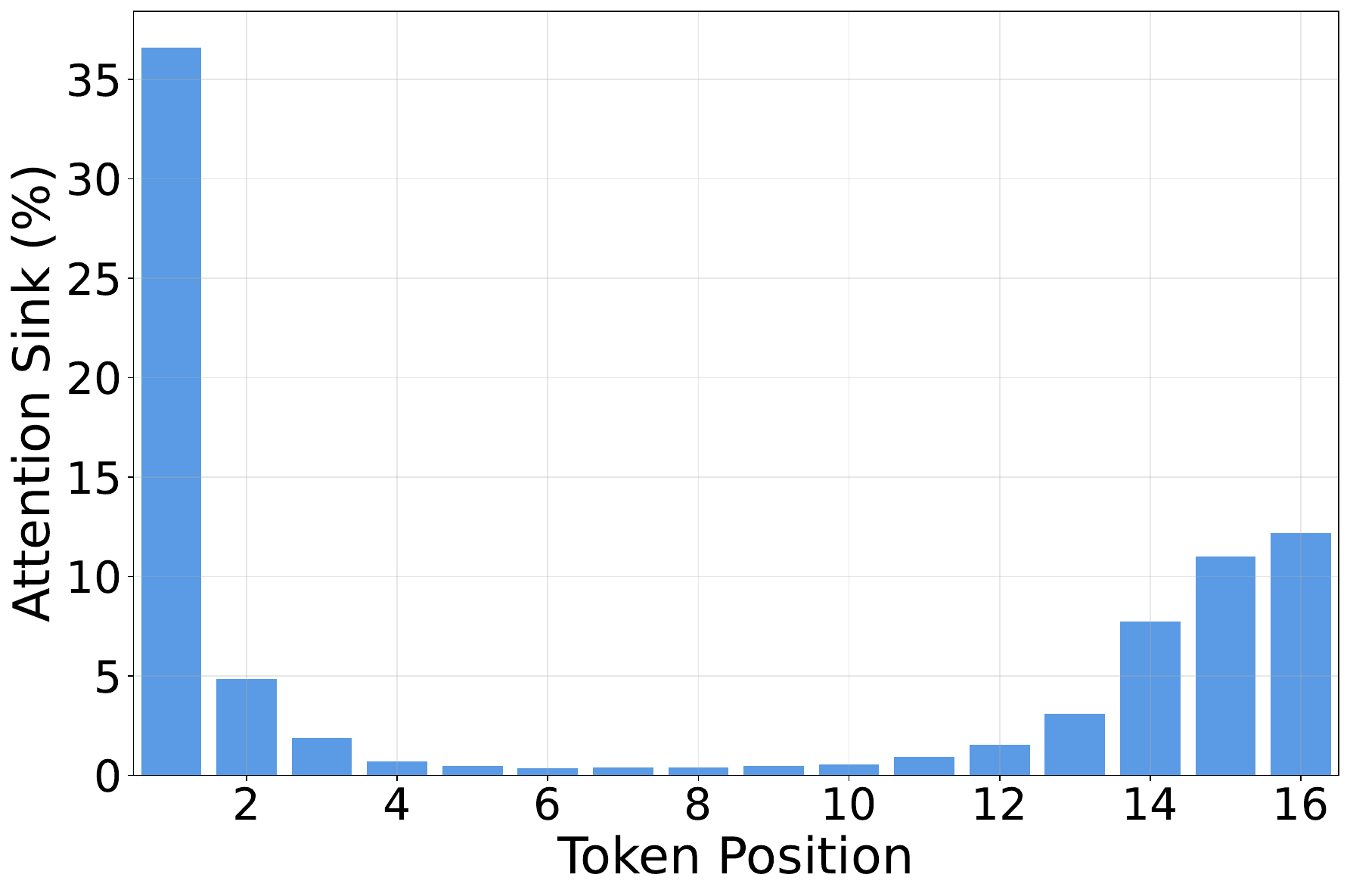}
        \caption{Attention sink (\%) across all token positions.}
        % \vspace{0.2cm}
        \label{fig:attn-sink-tokens}
    \end{subfigure}
    %\vspace{0.25cm}
    \caption{Attention sink score \citep{gu2025attention} analysis in the Falcon-Mamba-7B model. Subfigure (a) shows the sink score of the first token and its moving average across layers; the score steadily increases with depth and dominates in the final layers. Subfigure (b) shows the average sink scores across token positions; the first token consistently receives the largest share.}
    \label{fig:attn-sink-analysis}
\end{figure}

To quantify the attention sink in state-space models, we employ the metrics $\mathrm{Sink}_\ell$ and $\mathrm{Sink}$ (formally defined in Appendix~\ref{sec:attention-sink-appendix}), adapted from \citep{gu2025attention} with modifications to account for the structural differences in our hidden attention representation. Specifically, while the original metric is computed from attention logits with shape $(\text{batch}, \text{head}, \text{seqlen}, \text{seqlen})$, our Mamba hidden attention has shape $(\text{batch}, \text{channel}, \text{seqlen}, \text{seqlen})$. Each channel plays an analogous role to an attention head, but given the large number of channels ($4096$), we uniformly sample $200$ channels per layer for evaluation. We fix the threshold to $\varepsilon = 0.3$ to decide whether the first token exhibits sink in a given channel. The results, shown in Figure~\ref{fig:attn-sink-analysis}, reveal that the sink score of the first token increases consistently with depth and that, when averaged across token positions, the first token dominates all others, thereby extending the attention sink phenomenon to state-space models.

\section{Conclusion and Limitation}
%\vspace{-0.2cm}

In this work, we developed a theoretical framework for token dynamics in deep Mamba models under realistic assumptions. Modeling Mamba as flow maps on high-dimensional spheres, we characterized limit points of token trajectories, proved exponential convergence, and revealed the first token’s role as a global attractor. We further extended our analysis to time-dependent parameters setting and to broader autoregressive models, as well as demonstrated a practical implication to explain the attention sink effect.
%and a refinement via random token reordering. 
These results advance the understanding of token dynamics and suggest new directions for improving Mamba models.

\paragraph{Limitation.} Our analysis focuses on the continuous-time limit and asymptotic behavior, and future work is needed to fully bridge these insights with the discrete, large-scale implementations of Mamba models in practice.

% \newpage
% \section*{Impact Statement}

% This paper presents work whose goal is to advance the field of machine learning. There are many potential societal consequences of our work, none of which we feel must be specifically highlighted here.

% \newpage
\begin{ack}
This research is supported by the National Research Foundation Singapore under the AI Singapore Programme (AISG Award No: AISG2-TC-2023-012-SGIL). This research is also supported by the Ministry of Education, Singapore, under the Academic Research Fund Tier 1 (FY2023) (A-8002040-00-00, A-8002039-00-00), the NUS Presidential Young Professorship Award (A-0009807-01-00), the NUS Artificial Intelligence Institute--Seed Funding (A-8003062-00-00), and the Cross Faculty Grant 2025, CFG25-012 (A-8004460-00-00). T. T. Nguyen acknowledges support from the NSFC Excellent Young Scientists Fund (Overseas) and the Research Start-up Fund of Shanghai Jiao Tong University.
\end{ack}

\bibliography{icml2026_conference}
\bibliographystyle{plain}

%%%%%%%%%%%%%%%%%%%%%%%%%%%%%%%%%%%%%%%%%%%%%%%%%%%%%%%%%%%%%%%%%%%%%%%%%%%%%%%
%%%%%%%%%%%%%%%%%%%%%%%%%%%%%%%%%%%%%%%%%%%%%%%%%%%%%%%%%%%%%%%%%%%%%%%%%%%%%%%
% APPENDIX
%%%%%%%%%%%%%%%%%%%%%%%%%%%%%%%%%%%%%%%%%%%%%%%%%%%%%%%%%%%%%%%%%%%%%%%%%%%%%%%
%%%%%%%%%%%%%%%%%%%%%%%%%%%%%%%%%%%%%%%%%%%%%%%%%%%%%%%%%%%%%%%%%%%%%%%%%%%%%%%
\newpage
\appendix
\onecolumn

\begin{center}
{\bf \Large{Appendix of ``First-Token Attraction in Mamba Dynamics''}}
\end{center}

\DoToC

\section{Related Works}

\paragraph{ResNets and Neural ODEs}  
Our work connects to the literature that interprets deep neural networks (DNNs) through the lens of dynamical systems. Standard DNNs can be seen as discrete-time dynamical processes, where data propagate sequentially through layers \citep{lecun2015deep}. Residual networks (ResNets) \citep{chen2018neural, haber2017stable, kpotufe2012gradient} exemplify this view, and their continuous counterparts, neural ordinary differential equations (neural ODEs) \citep{lin2018resnet,zhang2020approximation,xia2021heavy,li2022deep,nguyen2022improvingnode,thorpe2022grand,ruiz2023neural, cheng2023interpolation}, formalize the passage to continuous time. In this formulation, the residual block update $z = x + y(x,\theta)$ becomes a flow governed by a differential equation, mapping an initial input $x(0)$ to an output $x(T)$. A broad line of work has studied the approximation, interpolation, and controllability properties of these architectures \citep{lin2018resnet, zhang2020approximation, li2022deep, tabuada2022universal, ruiz2023neural, cheng2023interpolation,teo2024momentumsmoe}.  

\paragraph{Oversmoothing and Rank Collapse}  
Transformers face two critical degeneracies: oversmoothing and rank collapse. Oversmoothing causes token representations to become indistinguishable, leading to a loss of expressive power. Nguyen et al.~\cite{nguyen2023mitigating,nguyen2024coupled}, Wu et al.~\cite{wu2024demystifying}, Dovonon et al.~\cite{dovonon2024setting}, and Chen et al.~\cite{chen2022principle} proved that self-attention alone cannot prevent this collapse. Remedies include \textit{ContraNorm}, a normalization method inspired by contrastive learning \citep{guo2023contranorm}, and the use of residual connections with normalization, which have been shown to mitigate oversmoothing \citep{scholkemper2024residual,abdullaev2025transformer,abdullaev2025revisiting}.  

Rank collapse, on the other hand, occurs when the hidden representations degenerate into a low-dimensional subspace. Prior studies \citep{dong2021attention, feng2022rank, noci2022signal, joudaki2023impact, zhai2023stabilizing,nguyen2024pidformer, noci2024shaped,nielsen2024elliptical, bao2024self} identified this as a major bottleneck and proposed stabilization strategies to maintain diversity and improve performance.

\paragraph{Theoretical Analyses of Mamba}  
Recent studies have begun to explore Mamba from a theoretical standpoint. Ali et al.~\cite{ali2024hidden} interpreted the S6 layer as a linear-time, softmax-free variant of attention, proving that Mamba is more expressive than transformers. In contrast, Muca Cirone et al.~\cite{cirone2025theoretical} applied Rough Path Theory to show that diagonal selective state space models (SSMs) are less expressive than non-diagonal ones. Merrill et al.~\cite{merrill2024illusion} used circuit complexity theory to argue that both SSMs and transformers belong to the class $TC^0$, implying computational equivalence. Jelassi et al.~\cite{jelassi2024repeat} provided both theoretical and empirical evidence that transformers surpass state space models in copying tasks. Together, these results form a theoretical basis for understanding the expressive and generalization limits of Mamba.

\paragraph{Token Dynamics and Clustering Effects}  
Another perspective interprets sequence models, such as Transformers and Mamba, as interacting particle systems. Lu et al.~\cite{lu2019understanding} formulated self-attention as a numerical solver for a convection--diffusion equation, while Dutta et al.~\cite{dutta2021redesigning} proposed \textit{TransEvolve}, a particle-inspired scheme that reduces complexity. These approaches connect Transformer dynamics to broader nonlinear systems, including flocking \citep{ha2019complete}, synchronization \citep{kuramoto1975self, acebron2005kuramoto}, and consensus models \citep{krause2000discrete, motsch2014heterophilious}.  

From this dynamical viewpoint, the long-term behavior of tokens has been studied via clustering. Geshkovski et al.~\cite{geshkovski2024emergence} proved convergence to limit objects influenced by initial conditions, while follow-up work \citep{geshkovski2023mathematical, geshkovski2024dynamic} showed metastable multi-cluster states that persist for long times before collapsing into a single cluster. Alcalde et al.~\cite{alcalde2024clustering} further confirmed asymptotic convergence to clustered equilibria in hardmax Transformers.  
Pham et al.~\cite{pham2025dynamical} investigated how different forms of positional encoding - specifically absolute and rotary - affect these dynamical regimes.
In contrast, the token dynamics of Mamba models remain largely unexplored. 
Vo et al.~\cite{vo2024demystifying} studied the dynamical properties of tokens in a deep Mamba model under a certain restrictive assumptions, and based on that, suggested improvements to the model by excluding negatively impactful scenarios and reordering tokens using importance scores.
% Existing analyses impose restrictive assumptions - such as one-dimensional embeddings or the absence of normalization layers - that oversimplify practical implementations~\cite{vo2024demystifying}. 

In this paper, we take one step further by studying the token dynamics for Mamba without the restrictive assumptions.

\section{Theoretical Results}

Recall that a pre-trained Mamba model can be viewed as a flow map on $(\mathbb{S}^{D-1})^L$ defined by the following dynamical system: 
\begin{align}\label{eq:mambaLN_dynamic_appendix}
    &\frac{d}{dt}x_{l}(t) = \mathbf{P}_{x_l(t)} \left( \sum\limits_{j=1}^l A_{lj}(\mathbf{x}(t)) x_{j}(t) \right), \quad t>0,\, l=1,\ldots,L,\\
    &\mathbf{x}(0)= (x_1(0),\ldots,x_L(0)) \in (\mathbb{S}^{d-1})^L,
\end{align}
where
    \begin{itemize}
        \item for each $x \in \mathbb{S}^{D-1}$, the map $\mathbf{P}_{x} \colon \mathbb{R}^D \to T_{x}\mathbb{S}^{D-1}$ is the projection onto the tangent space at $x$ of the sphere $\mathbb{S}^{D-1}$ and it is defined by
        $$\mathbf{P}_x(y) = y - \left< y,x\right>x, \quad y \in \mathbb{R}^D;$$

        \item the coefficient matrices $A_{lj}$ is an $D \times D$ diagonal matrix $A_{lj} = \operatorname{diag}(A_{1,lj},\ldots,A_{D,lj})$ with
        \begin{equation}
            A_{d,lj}(\mathbf{x}(t)) = \left\{
            \begin{aligned}
                & x_l(t)^{\top} S_{BC} x_l(t) \cdot \Delta_d(x_l(t)), &\text{if } l = j,\\
                &x_l(t)^{\top} S_{BC} x_j \cdot \Delta_d(x_j(t)) \cdot \exp \left( -a_d(t) \sum\limits_{k=j+1}^l \Delta_d(x_k(t)) \right), &\text{if } l > j;
            \end{aligned}
            \right.
        \end{equation}

        \item the step size function at the $d$-th channel $\Delta_d \,:\, \mathbb{R}^D \to \mathbb{R}$ is defined by 
        \begin{align}
            \Delta_d(u) = \text{softplus}(S_{\Delta,d}(t) \cdot u) = \ln \left( 1+e^{S_{\Delta, d} \cdot u} \right), \qquad u \in \mathbb{R}^D.
        \end{align}

    \end{itemize}

    The dynamical system~\eqref{eq:mambaLN_dynamic} is parameterized by the input-output matrix 
    \[
        S_{BC} = S_C^{\top} S_B \in \mathbb{R}^{D \times D},
    \]
    the step-size matrix 
    \[
        S_{\Delta} = [S_{\Delta,1}, \ldots, S_{\Delta,D}] \in \mathbb{R}^{D \times D},
    \]
    and the state coefficient vector 
    \[
        a= (a_1, \ldots, a_D) \in \mathbb{R}_{>0}^{D}.
    \]

Our main problem is studying the dynamical properties of the solutions of system~\eqref{eq:mambaLN_dynamic}.
In particular, we want to understand how the set $\lim_{t \to +\infty} \{x_1(t),\ldots,x_L(t)\}$ looks like and when it reduces to a set of a single point.

\paragraph{Notation.} Throughout this section, the notation $O(1)$ denotes a universal positive constant. We write $A \lesssim B$ to mean that there exists a constant $C_0 > 0$, independent of time, such that $|A| \leq C_0 |B|$.

\subsection{Possible Limit Points of the First Token}

In this subsection, we will characterize all possible limit points for the first token.
In order to do that, let us consider the differential equation governing the trajectory of the first equation, which is the first equation in the dynamical system~\eqref{eq:mambaLN_dynamic_appendix}, given by
\begin{equation}\label{eq:first_token_appendix}
    \frac{d}{dt} x_1 = \mathbf{P}_{x_1}\big(A_{11}(x) x_1\big)
\end{equation}
where
\[
A_{11}(x) = \operatorname{diag}(A_{1,11}, A_{2,11}, \cdots, A_{d,11})
\]
and $A_{d,11} = \langle x_1, S_{BC} x_1 \rangle \Delta_d(x_1)$. 
We will determine the set of equilibrium points of this differential equation in Subsection~\ref{sec:first_token}.
However not every equilibrium point is a limit point.
The set of possible actual limit points of the first token is then presented in Subsection~\ref{sec:first_token_stability}.
Moreover, since 
\[
\langle S_{BC}x_1, x_1 \rangle = \langle Sx_1, x_1 \rangle,
\]
where 
\[
S = \tfrac{1}{2}\bigl(S_{BC} + S_{BC}^\top\bigr)
\]
denotes the symmetric part of $S_{BC}$, it follows that only the symmetric component contributes to the quadratic form.  
Throughout this section, we may therefore assume without loss of generality that $S_{BC}$ is symmetric.

\subsubsection{Equilibrium Points According to the First Token}
\label{sec:first_token}
\begin{theorem}[Equilibria of the first token]\label{thm:x1*}
A vector $y \in \mathbb{S}^{D-1}$ is an equilibrium of equation~\eqref{eq:first_token_appendix} 
if and only if it belongs to one of the following types:
\begin{itemize}
    \item[\textbf{Type 1.}] (Alignment equilibria)  
    The vector $y$ may have some zero coordinates, but restricted to its nonzero entries we obtain
    \[
        y^{(\ne 0)} = (y_{k_1},\dots, y_{k_\ell}) \in \mathbb{S}^{\ell-1},
    \]
    which lies in the common kernel of the column differences
    \[
        S_{\Delta,k_j} - S_{\Delta,k_1}, \qquad j=1,\dots,\ell,
    \]
    for the active indices $\{k_1,\dots,k_\ell\}$.

    \item[\textbf{Type 2.}] (Quadratic constraint equilibria)  
    The vector $y$ satisfies
    \[
        y^{\top} S_{BC}\, y = 0.
    \]
\end{itemize}

If $S_{BC}$ is definite, then Type~2 equilibria do not exist.  
Furthermore, if the columns of $S_{\Delta}$ are linearly independent, then the Type~1 equilibria collapse to exactly two antipodal points
\[
y^{(\neq 0)} = \pm \xi,
\]
where $\xi$ is the unique unit vector in the one-dimensional intersection of the kernels above.
\end{theorem}

\begin{proof}
By definition,
\[
A_{11}(x) = \langle x_1, S_{BC} x_1 \rangle 
\operatorname{diag}\bigl(\Delta_1(x_1), \Delta_2(x_1), \ldots, \Delta_D(x_1)\bigr).
\]
Substituting $x_1 = y = (y_1,\dots,y_D) \in \mathbb{S}^{D-1}$, the dynamics take the form
\[
y' = \mathbf{P}_y\bigl(A_{11}(y)y\bigr) 
= \langle S_{BC} y, y \rangle \, \mathbf{P}_y\bigl(\Delta_1(y)y_1,\ldots,\Delta_D(y)y_D\bigr),
\]
where $\mathbf{P}_y$ denotes the orthogonal projection onto $T_y\mathbb{S}^{D-1}$.  

For each component this yields
\begin{equation}\label{eq:y_i}
    y_i' = \langle S_{BC} y, y \rangle \,\bigl\{\Delta_i(y) - \Delta(y)\bigr\} y_i,
\end{equation}
with
\begin{equation}\label{eq:Delta}
    \Delta(y) = \sum_{k=1}^D \Delta_k(y) y_k^2, 
    \qquad 
    \Delta_i(y) = \ln\bigl(1+e^{S_{\Delta,i}\cdot y}\bigr).
\end{equation}

Equation \eqref{eq:y_i} shows that equilibria arise in two distinct cases:
\begin{itemize}
    \item[(i)] either $\langle S_{BC} y, y \rangle = 0$, corresponding to Type~2 equilibria,  
    \item[(ii)] or else $\Delta_i(y) = \Delta(y)$ for every index $i$ with $y_i \neq 0$.
\end{itemize}

Since $|y|=1$, there exists at least one index $i_0$ with $y_{i_0}\neq 0$.  
For any $i$ with $y_i \neq 0$, condition (ii) implies
\[
\Delta_i(y) = \Delta_{i_0}(y),
\]
which is equivalent to
\begin{equation}\label{S-perp}
(S_{\Delta,i}-S_{\Delta,i_0})\cdot y = 0.
\end{equation}

Let $\{k_1,\dots,k_\ell\}$ denote the indices of the nonzero entries of $y$, and write 
\[
y^{(\neq 0)}=(y_{k_1},\dots,y_{k_\ell})
\]
for the restriction of $y$ to these coordinates.  
Then \eqref{S-perp} becomes
\[
(S_{\Delta,k_j} - S_{\Delta,k_1}) \cdot y^{(\neq 0)} = 0,
\qquad j=1,\dots,\ell,
\]
which precisely characterizes Type~1 equilibria.  

Finally, if $S_{BC}$ is definite, then $\langle S_{BC} y, y \rangle > 0$ for all $y \neq 0$, so no Type~2 equilibria can exist.  
Moreover, if the columns $\{S_{\Delta,k_j}\}_{j=1}^\ell$ are linearly independent, the intersection
\[
\bigcap_{j=1}^\ell \langle S_{\Delta,k_j} - S_{\Delta,k_1}\rangle^{\perp}
\]
is one-dimensional in $\mathbb R^\ell$.  
Its intersection with the unit sphere $\mathbb{S}^{\ell-1}$ therefore consists of exactly two antipodal points $\pm \xi$, which are the only possible equilibria.
\end{proof}

% In what follows, we study the stability of the equilibrium points.

\subsubsection{Limit points of the First Token}
\label{sec:first_token_stability}

The analysis of stability near equilibrium points, in order to characterize the limit sets of
equation~\eqref{eq:first_token_appendix}, is in general quite delicate.  
Thanks to Theorem~\ref{thm:x1*}, under the assumption that $S_{BC}$ is definite and the columns of
$S_{\Delta}$ are linearly independent, the nonzero-part of equilibria are the two antipodal points
\[
y^{(\neq 0)} =(y_{k_1},\cdots, y_{k_\ell})= \pm \xi,
\]
where $\xi$ is the unique intersection of $\mathbb{S}^{\ell-1}$ with the one-dimensional subspace
\[
\bigcap_{j=1}^{\ell}\ker\!\bigl((S_{\Delta,k_j}-S_{\Delta,k_1})^{\top}\bigr).
\]

To avoid unnecessary technical complications, we shall henceforth restrict attention to this case.
In particular, we consider the high-dimensional setting in which
\[
S_{\Delta,m} = \pm e_m,
\]
with $e_m$ denoting the $m$-th coordinate vector (see Theorem~\ref{thm:first_token}).  
Apart from the special case $D=2$, where the structure can be worked out explicitly
(see Theorem~\ref{thm:first_token_D=2}), this reduction allows us to describe all possible limit
points of the first token.

We start with the following auxiliary lemma.
\begin{lemma}[Sign Preservation]\label{lem:yi_positive}
Let $y(t)=(y_1(t),\ldots,y_D(t)) \in C^{\infty}([0,+\infty),\mathbb{R}^D)$ be a solution of equation~\eqref{eq:first_token_appendix}.
If $y_i(0) < 0$ then $y_i(t) < 0$ for all $t > 0$.  
Similarly, if $y_i(0) > 0$ then $y_i(t) > 0$ for all $t > 0$.
\end{lemma}

\begin{proof}
Suppose $y_i(0) < 0$ but $y_i(t) \geq 0$ for some $t > 0$.  
Then there exists a minimal $T > 0$ such that
\[
    y_i(T) = 0, \qquad y_i(t) < 0 \quad \text{for all } t \in [0,T).
\]
By continuity of $y_i$, we have
\[
    \lim_{t \to T^-} y_i(t) = 0.
\]

For $t \in (0,T)$ the evolution equation reads
\[
    \frac{d}{dt} y_i(t) = (S_{BC}y \cdot y)\bigl(\Delta_i(y) - \Delta(y)\bigr)\, y_i(t).
\]
Dividing by $y_i(t)$ yields
\[
    \frac{d}{dt}\bigl(\ln |y_i(t)|\bigr)
    = (S_{BC}y \cdot y)\bigl(\Delta_i(y) - \Delta(y)\bigr).
\]

Let $\{t_n\}_{n=1}^\infty \subset (0,T)$ be an increasing sequence with $t_n \to T$.  
Then
\[
    \ln |y_i(t_n)| - \ln |y_i(0)|
    = \int_0^{t_n} (S_{BC}y \cdot y)\bigl(\Delta_i(y) - \Delta(y)\bigr)\,ds.
\]
The right-hand side is bounded, hence the left-hand side is also bounded.  
But since $|y_i(t_n)| \to 0$, we have $\ln |y_i(t_n)| \to -\infty$, a contradiction.  
Thus $y_i(t) < 0$ for all $t > 0$.  

The case $y_i(0) > 0$ is analogous.
\end{proof}

\begin{lemma}\label{lem:lim_yi_yj}
	Let $y(t)=(y_1(t),\ldots,y_D(t)) \in C^{\infty}([0,+\infty),\mathbb{R}^D)$ be a solution of equation~\eqref{eq:first_token_appendix}.
	Assume that $S_{\Delta,m} = \pm e_m$ and $y^{\top}S_{BC}y \geq  0$ for all $y \neq 0$. 
	Here, the sign $\pm$ in $\pm e_m$ is the sign of $y_m(0)$. Suppose that there are $i, j\in\{1,..., D\}$ such that
	$$0<|y_j(0)|<|y_i(0)|.$$
	Then 
	$$0<|y_j(t)|<|y_i(t)|,$$
	for every $t>0$.
\end{lemma}

\begin{proof}
	By Lemma \ref{lem:yi_positive}, we have $y_j(t), y_i(t)\neq 0$ for every $t>0$. It remains to prove that
	$$|y_j(t)|<|y_i(t)|,$$
	for every $t>0$.
	
	The same computation as in the proof of Lemma \ref{lem:yi_positive} gives
	$$\frac{d}{dt}\bigl(\ln |y_i(t)|\bigr)
	= (S_{BC}y \cdot y)\bigl(\Delta_i(y) - \Delta(y)\bigr),$$
	and
	$$\frac{d}{dt}\left(\ln |y_i(t)|\right)
	= (S_{BC}y \cdot y)\bigl(\Delta_i(y) - \Delta(y)\bigr).$$
	Hence, we have
	\begin{equation}\label{eq1:lem:lim_yi_yj}
		\frac{d}{dt}\bigl(\ln \frac{|y_i(t)|}{|y_j(t)|}\bigr)=(S_{BC}y \cdot y)\bigl(\Delta_i(y) - \Delta_j(y)\bigr).
	\end{equation}
	Note that, since $y_j$ and $y_i$ are continuous and nonvanishing on $[0, \infty)$, they have constant sign on $[0, \infty)$.
	Therefore,
	\begin{center}
		$S_{\Delta, i}(y(t))=\textnormal{sign}(y_i(0))y_i(t)=|y_i(t)|$
		and
		$S_{\Delta, j}(y(t))=\textnormal{sign}(y_j(0))y_j(t)=|y_j(t)|$,
	\end{center} 
	for every $t>0$. Consequently, we get
	\begin{center}
		$\Delta_d(y(t))=\ln\left(1+e^{|y_d(t)|}\right),$
	\end{center}
	for $d=i, j$. Combining this with \eqref{eq1:lem:lim_yi_yj}, we obtain
	\begin{equation}\label{eq2:lem:lim_yi_yj}
		\frac{d}{dt}\bigl(\ln \frac{|y_i(t)|}{|y_j(t)|}\bigr)=(S_{BC}y \cdot y)\left(\ln\left(1+e^{|y_i(t)|}\right)
		-\ln\left(1+e^{|y_j(t)|}\right)\right).	
	\end{equation}
	Now, suppose to the contrary that $|y_j(t)|\geq |y_i(t)|$ for some $t>0$. Then there is $T>0$ such that 
 $|y_j(t)|< |y_i(t)|$ for every $0\leq t<T$ and  $|y_j(T)|=|y_i(T)|$. Integrating both side of \eqref{eq1:lem:lim_yi_yj} and using the
 fact 	$S_{BC}y \cdot y\geq 0$, we get
 $$\ln \frac{|y_i(T)|}{|y_j(T)|}-\ln \frac{|y_i(0)|}{|y_j(0)|}=\int_0^T(S_{BC}y \cdot y)\left(\ln\left(1+e^{|y_i(t)|}\right)
 -\ln\left(1+e^{|y_j(t)|}\right)\right)ds\geq 0.$$
 Then
 $$0=\ln\frac{|y_i(T)|}{|y_j(T)|}\geq \ln \frac{|y_i(0)|}{|y_j(0)|}>0.$$
 We get a contradiction. Thus, $|y_j(t)|<|y_i(t)|$ for every $t>0$.
 
 The proof is complete.
\end{proof}

\begin{lemma}\label{lem:lim_yi_yj_cont}
Let $y(t) = (y_1(t),\ldots,y_D(t)) \in C^{\infty}([0,+\infty),\mathbb{R}^D)$ be a solution of equation~\eqref{eq:first_token_appendix}. 
Assume that the symmetric part 
\[
S = \tfrac{1}{2}(S_{BC} + S_{BC}^{\top})
\]
of $S_{BC}$ is positive definite, and that $S_{\Delta,m} = \pm e_m$, where the sign $\pm$ agrees with the sign of $y_m(0)$. 
Suppose that there are $i, j\in\{1,..., D\}$ such that
$$0<|y_j(0)|<|y_i(0)|.$$
Then
$$\lim_{t\to\infty}y_j(t)=0.$$
\end{lemma}

\begin{proof}
Define
\[
c(t) = \frac{|y_i(t)|}{|y_j(t)|}.
\]
By Lemma~\ref{lem:lim_yi_yj}, we have $c(t) > 1$ for all $t \geq 0$.  
Moreover, it follows from \eqref{eq2:lem:lim_yi_yj} that
	\begin{equation}\label{eq1:lem:lim_yi_yj_cont}
	\frac{d}{dt}\ln c(t)=(S y \cdot y)\left(\ln\left(1+e^{|y_i(t)|}\right)
	-\ln\left(1+e^{|y_j(t)|}\right)\right)\geq 0.
\end{equation}
In particular, $c(t)$ is increasing. 

We distinguish two cases.

{\bf Case 1: } $c(t)\rightarrow\infty$ as $t\rightarrow\infty$.\\
In this case we have 
\begin{center}
	$|y_j(t)|\leq \frac{1}{c(t)}\rightarrow 0$ as $t\rightarrow\infty$.
\end{center}
{\bf Case 2: } $c(t)$ is bounded on $(0, \infty)$.\\
Let $f(t) = \ln(1+e^t)$. By the mean value theorem, there exists $u \in (|y_j|, |y_i|)$ such that
\[
f(|y_i|) - f(|y_j|) = f'(u)\,(|y_i| - |y_j|) 
= \frac{e^u}{1+e^u}\, (c(t)-1)|y_j(t)|\geq \frac{(c(0)-1)|y_j(t)|}{2}.
\]
Combining this with \eqref{eq1:lem:lim_yi_yj_cont} gives
\begin{equation}\label{eq2:lem:lim_yi_yj_cont}
	|y_j(t)|\leq C\,\frac{d}{dt}\ln c(t),
\end{equation}
where $C=\frac{2}{(c(0)-1)\min_{y\in\mathbb{S}^{D-1}}(Sy\cdot y)}>0$. Integrating both sides of \eqref{eq2:lem:lim_yi_yj_cont}, we get
	$$\int_0^{\infty}|y_j(t)|dt\leq C(\lim_{T\to\infty}\ln c(T)-\ln c(0))<\infty.$$
	As a consequence, we have
	$$\lim_{n\to\infty}\int_n^{n+1}|y_j(t)|dt=0.$$
	Since $y_j(t)$ is Lipschitz continuous, it follows that 
	$$\lim_{t\to\infty}|y_j(t)|=0.$$
This completes the proof.
\end{proof}

\begin{theorem}[Convergence of the first tokens in high dimension]\label{thm:first_token}
Assume that the symmetric part 
of $S_{BC}$ is positive definite.
    Let $y(t) = (y_1(t),\ldots,y_D(t)) \in C^{\infty}([0,\infty),\mathbb{R}^D)$ be a solution of equation~\eqref{eq:first_token_appendix}.
    Assume that $S_{\Delta,m} = \pm e_m$, where the sign is chosen according to $\operatorname{sgn}(y_m(0)) \neq 0$, for all $m=1,\ldots,D$.
    Suppose that there exists $d_0\in\{1,..., D\}$ such that
    $$|y_{d_0}(0)|>|y_d(0)|>0,$$
    for every $d\neq d_0$.
    Then
    \[
        \lim_{t \to \infty} y(t) = \pm e_{d_0},
    \]
    where the sign is $\textnormal{sign}(y_{d_0}(0))$.
\end{theorem}

\begin{proof}
 By Lemma \ref{lem:lim_yi_yj_cont}, we have
 \begin{equation}\label{eq1:thm:first_token}
 \lim_{t\to\infty}y_d(t)=0,	
 \end{equation}
 for every $d\neq d_0$. Then
 $$\lim_{t\to\infty}|y_{d_0}(t)|=1.$$
 Since $y_{d_0}(t)$ has constant sign on $[0, \infty)$ , it follows that
 \begin{equation}\label{eq2:thm:first_token}
 	\lim_{t\to\infty}y_{d_0}(t)=\textnormal{sign}(y_{d_0}(0)).	
 \end{equation}
Combining \eqref{eq1:thm:first_token} and \eqref{eq2:thm:first_token}, we get
$$ \lim_{t \to \infty} y(t) = \textnormal{sign}(y_{d_0}(0)) e_{d_0},$$
as desired.
\end{proof}

\begin{remark}
    The conclusion of Theorem~\ref{thm:first_token} highlights a striking selection mechanism:
    under the assumption that the columns of $S_\Delta$ are linearly independent, the long-time
    dynamics do not converge to a large collection of equilibria lying in a lower-dimensional sphere in Theorem \ref{thm:x1*}.
    Instead, the trajectory is forced to pick out one of the most ``extreme'' coordinate directions,
    namely the basis vector $\pm e_d$ corresponding to the largest initial coordinate in magnitude. 

\end{remark}

In the two-dimensional case, we can actually determine all possible limit points of the first token in the following theorem.
\begin{theorem}[Convergence of the first token in two dimensions]\label{thm:first_token_D=2}
    Let $D=2$ and let $y(t) = (y_1(t),y_2(t)) \in C^{\infty}([0,+\infty),\mathbb{S}^1)$ be a solution of~\eqref{eq:first_token_appendix}.
    Assume that
    \begin{itemize}
        \item[$\bullet$] $S_{BC} = \operatorname{diag}(\lambda_1,\lambda_2)$ with either $\lambda_1 \neq \lambda_2$ or $\lambda_1 = \lambda_2 \neq 0$;
        \item[$\bullet$] $S_{\Delta,2} \neq S_{\Delta,1}$. 
    \end{itemize}
    In this setting, the unit circle $\mathbb{S}^1$ intersects the line 
    $\langle S_{\Delta,2} - S_{\Delta,1}\rangle^{\perp}$ at two antipodal points $\pm \xi$.
    Then the limit $\lim_{t \to +\infty} y(t)$ exists and belongs to the set
    \[
        \Omega 
        = \{(0,\pm 1),(\pm 1,0)\} 
        \cup \{\pm \xi\} 
        \cup \left\{ (\cos\varphi_\star,\sin\varphi_\star) : \lambda_1\cos^2\varphi_\star + \lambda_2\sin^2\varphi_\star = 0 \right\}.
    \]
\end{theorem}
\begin{proof}
    Since $D=2$, we may parametrize
    \[
        y(t) = (\cos \varphi(t), \, \sin \varphi(t)), \qquad \varphi(t)\in[0,2\pi].
    \]
    Define $y^{\perp} = (-y_2,y_1)$. Then
    \[
        \varphi'(t) y^{\perp} = y'(t) = A_{11}y - (A_{11}y \cdot y)\,y.
    \]
    Taking the inner product with $y^{\perp}$, and using $|y|=1$ and $y\cdot y^{\perp}=0$, yields
    \[
        \varphi'(t) = A_{11}y \cdot y^{\perp} 
        = (\lambda_1 y_1^2 + \lambda_2 y_2^2) y_1 y_2 \bigl(\Delta_2(y) - \Delta_1(y)\bigr).
    \]
    Substituting $y=(\cos\varphi,\sin\varphi)$, we obtain
    \[
        \varphi'(t) = \bigl[(\lambda_1+\lambda_2) + (\lambda_1-\lambda_2)\cos(2\varphi)\bigr] 
        \sin(2\varphi)\,
        \bigl[h(R_2,\varphi-\beta) - h(R_1,\varphi-\alpha)\bigr],
    \]
    where
    \[
        h(R,u) = \tfrac{1}{4}\ln\!\bigl(1+e^{R\cos u}\bigr), 
        \qquad S_{\Delta,2} = R_2(\cos\beta,\sin\beta), 
        \qquad S_{\Delta,1} = R_1(\cos\alpha,\sin\alpha).
    \]
    
    Thus $\varphi$ satisfies the autonomous ODE
    \[
        \varphi' = F(\varphi),
    \]
    where the points $(\cos\varphi,\sin\varphi)$ corresponding to zeros of $F$ in $[0,2\pi]$ form precisely $\Omega$. Since $\Omega$ is finite, the zeros of $F$ partition $(0,2\pi)$ into intervals on each of which $F$ has a definite sign. By the phase portrait of $\varphi$, every trajectory converges to one of the equilibria of $F$, and hence $y(t)$ converges to a point in $\Omega$. This proves the claim.
\end{proof}

\subsection{All Other Tokens are Attracted by the First Token}\label{sec:appendix_all_tokens}

Different from the first token, the study of the convergence of all tokens based on the dynamical system~\eqref{eq:mambaLN_dynamic_appendix} is much more challenging. 
To make the proofs feasible while maintaining the generality of our theoretical findings, in this whole subsection, we will consider the dynamical system~\eqref{eq:mambaLN_dynamic_appendix} under the following assumptions:
\begin{itemize}
    
    \item[]  \textit{Assumption:} $S_{\Delta,1} = \ldots = S_{\Delta,D} =: \alpha \in \mathbb{R}^D$ and $a_1=\ldots=a_D =: a \in \mathbb{R}$.

\end{itemize}

In this case we can rewrite dynamical system~\eqref{eq:mambaLN_dynamic_appendix} as:
\beq\label{new-eq}
    \frac{d}{dt} x_i=\sum_{j=1}^{i-1}A_{ij}\mathbf{P}_{x_i}(x_j)
\eeq
where $A_{ij}$ is defined as
\[
A_{ij}=(x_i\cdot S_{BC}x_j)a_{ij},\qquad a_{ij}=\triangle(x_i)e^{-a\sum_{k=j+1}^i \triangle(x_k)}, \qquad
\text{and} \qquad 
\triangle(x_i)=\ln(1+e^{\alpha \cdot x_i}).
\]

Throughout this section, we will assume that $x(t)=(x_1(t),x_2(t),\ldots,x_L(t)) \in C^{\infty}([0,+\infty);\mathbb{R}^D)^L$ is a solution of the dynamical system~\eqref{new-eq}.

For the first token $x_1(t)$, we see that 
\[
      \frac{d}{dt} x_1=A_{11}\mathbf{P}_{x_1}(x_1)=0
\]
This implies that $x_1(t)=x_1$ is a constant for all $t\ge 0$.
In the following, we will present all possible limit points of $x_l(t)$ with $l=2,\ldots,L$.

% \subsubsection{All Possible Limit Points}\label{sec:all_limit_points_appendix}

Define the consensus set
\[
\Omega_0 = \{(\pm x_1,\ldots,\pm x_1)(0)\in(\mathbb{S}^{D-1})^{L-1}\}.
\]
This set has exactly $2^{L-1}$ distinct elements, corresponding to the situation where every token is either aligned or anti-aligned with $x_1$.

Our main theorem of this section is stated as follows:

\begin{theorem}\label{all-conv}
Consider $L$ trajectories $x_1(t),\ldots,x_L(t)$ evolving on the unit sphere $\mathbb{S}^{D-1}$ under the dynamics~\eqref{new-eq}. 
Assume that $S_{BC}$ is a $D\times D$ symmetric matrix, and that the initial reference token $x_1=x_1(0)$ satisfies
\[
\langle x_1, S_{BC} x_1 \rangle > 0.
\]
Suppose further that for each $i\ge 2$,
\[
\langle x_i(0), S_{BC} x_1 \rangle \notin \{0,-1,1\},
\qquad 
\langle x_i(0), x_1 \rangle \notin \{-1,1\}.
\]
Then there exists a set of measure zero $N \subset (\mathbb{S}^{D-1})^{L-1}$, depending only on $x_1$ and $S_{BC}x_1$, such that for any initialization $(x_2,\ldots,x_L)(0)\in (\mathbb{S}^{D-1})^{L-1}\setminus N$, precisely one of the trajectory $x(t)=(x_2,\ldots,x_L)(t)$ converges to some $\omega_0\in\Omega_0$ as $t\to\infty$. Moreover, there exists $C_0>0$ such that
\[
\|x(t)-\omega_0\|\lesssim e^{-C_0 t}, \qquad t\ge 0.
\]
\end{theorem}

\paragraph{Outline of the proof.}
The structure of this section is as follows. We first establish an affirmative result: under a favorable positive-angle condition on the initial token configuration, the system converges to a state where all tokens align with $x_1$. This is proved in Proposition~\ref{conv-to-x1}, with the base case $L = 2$ treated separately in Proposition~\ref{2ndtoken}. We then proceed by induction: assuming exponential convergence of the first $k-1$ tokens to $\pm x_1$, we show that the $k$-th token either converges exponentially to $\pm x_1$ or to a point in the orthogonal complement $\langle S_{BC} x_1 \rangle^\perp$. Finally, Proposition~\ref{zero-prop} demonstrates that the latter scenario occurs only for a set of initial conditions of measure zero. The main difficulty arises from the fact that the interaction terms $\langle x_i, S_{BC} x_j \rangle$ are not sign-definite---a feature that distinguishes the model from classical consensus or synchronization dynamics. This necessitates a careful combination of nonlinear analysis and linear stability techniques to exclude nongeneric limiting behaviors.
\begin{proposition}\label{conv-to-x1}
Assume that $\langle x_1(0), S_{BC} x_1(0) \rangle > 0$ and that:
\begin{itemize}
    \item $\langle x_i(0), S_{BC} x_j(0) \rangle > 0$ for all $i, j \in [1, L]$, and
    \item $\langle x_i(0), x_1(0) \rangle \notin \{-1, 1\}$ for all $i \in [2, L]$.
\end{itemize}
Then,
\[
\sum_{i=1}^L |x_i(t) - x_1(0)| \lesssim e^{-O(1)t}.
\]
Moreover, we have
\[
\langle x_i(t), S_{BC} x_j(t) \rangle > 0 \quad \text{for all } t > 0.
\]
\end{proposition}

The proof of Proposition~\ref{conv-to-x1} will be postponed to the end of this section, as it requires several preparatory steps.  
To analyze the token dynamics, we first encode the pairwise interactions between tokens through the scalar quantities
\[
X_{ij}(t) = \la x_i(t), x_j(t)\ra, 
\qquad 
U_{ij}(t) = \la x_i(t), S_{BC}x_j(t)\ra,
\qquad i,j\in [1,L].
\]
Here $X_{ij}$ measures the relative alignment between tokens $i$ and $j$ on the sphere $\S^{d-1}$, while $U_{ij}$ captures their interaction through the transformation $S_{BC}$. 
These variables will serve as the fundamental building blocks in our nonlinear analysis and will play a central role in the proof of Theorem~\ref{all-conv}.  

Our first step is to establish their evolution equations.

\begin{lemma}\label{XU}
Assume that $S_{BC}$ is symmetric. 
For $i>1$, we have
\beq
\bega
X_{i1}'(t)
&=\sum_{j=1}^{i-1}a_{ij}U_{ij}\lw[ X_{j1}-X_{ij}X_{i1}\rw],\\
U'_{i1}(t)
&=\sum_{j=1}^{i-1}a_{ij}U_{ij}\lw[U_{j1}-X_{ij}U_{i1}\rw].
\enda
\eeq
If $i\ge j>1$, then 
\[
\bega
X'_{ij}(t)
&=\sum_{k=1}^{i-1}a_{ik}U_{ik}\lw[X_{kj}-X_{ik}X_{ij}\rw]
+\sum_{k=1}^{j-1}a_{jk}U_{jk}\lw[X_{ki}-X_{jk}X_{ji}\rw],\\
U'_{ij}(t)
&=\sum_{k=1}^{i-1}a_{ik}U_{ik}\lw[U_{kj}-X_{ik}U_{ij}\rw]
+\sum_{k=1}^{j-1}a_{jk}U_{jk}\lw[U_{ki}-X_{jk}U_{ji}\rw].
\enda
\]
\end{lemma}
\begin{proof}
Recalling \eqref{new-eq}, we have
\[
      \frac{d}{dt} x_i=\sum_{j=1}^{i-1}a_{ij}U_{ij}\lw[x_j-X_{ij}x_i
\rw].
\]
Dotting both sides of the above equation with $x_1$, we have 
\[\bega
X_{i1}'(t)
&=\sum_{j=1}^{i-1}a_{ij}U_{ij}\lw[ X_{j1}-X_{ij}X_{i1}
\rw].
\enda
\]
Next for $i\ge j>1$, we compute
\[\bega 
X'_{ij}&=\frac{d}{dt}\la x_i(t),x_j(t)\ra=      \frac{d}{dt} x_i\cdot x_j+      \frac{d}{dt} x_j\cdot x_i\\
&=\sum_{k=1}^{i-1}a_{ik}U_{ik}(x_k\cdot x_j-X_{ik}x_i\cdot x_j)+\sum_{k=1}^{j-1}a_{jk}U_{jk}(x_k\cdot x_i-X_{jk}x_j\cdot x_i)\\
&=\sum_{k=1}^{i-1}a_{ik}U_{ik}(X_{kj}-X_{ik}X_{ij})+\sum_{k=1}^{j-1}a_{jk}U_{jk}(X_{ki}-X_{jk}X_{ji}).
\enda
\]
We also have
\[
\frac{d}{dt}U_{i1}=      \frac{d}{dt} x_i\cdot S_{BC}x_1=\sum_{j=1}^{i-1}a_{ij}U_{ij}\lw[U_{j1}-X_{ij}U_{i1}
\rw],
\]
and similarly
\[\bega 
      \frac{d}{dt} U_{ij}&=      \frac{d}{dt} x_i\cdot S_{BC}x_j+      \frac{d}{dt} x_j\cdot S_{BC}x_i\\
&=\sum_{k=1}^{i-1}a_{ik}U_{ik}\lw[U_{kj}-X_{ik}U_{ij}
\rw]+\sum_{k=1}^{j-1}a_{jk}U_{jk}\lw[U_{ki}-X_{jk}U_{ji}
\rw].
\enda 
\]
The proof is complete.
\end{proof}
\begin{proposition}\label{limit}
Assume that $S_{BC}$ is symmetric and that the leading token $x_1=x_1(0)$ satisfies
\[
\langle x_i(0),\, S_{BC} x_1 \rangle \notin \{-1,0,1\},
\qquad\text{for all }\, i\in[2,L].
\]
Then there exists a minimal index $k\in[3,L]$ such that exactly one of the following holds:

\begin{itemize}

\item[(i)]
Every token up to level $k$ converges exponentially to the expected limit direction:
\[
x_i(t)\to \pm x_1(0)
\quad\text{as}\quad t\to\infty,\qquad i\le k,
\]
with uniform exponential rate
\[
\sum_{i=1}^k \bigl|x_i(t)-(\pm x_1(0))\bigr|
\;\lesssim\; e^{-O(1)t}.
\]

\item[(ii)] All earlier tokens align exponentially:
\[
x_i(t)\to \pm x_1(0),\qquad i\le k-1,
\]
while the $k$-th token loses its component along the direction $S_{BC}x_1(0)$:
\beq\label{exp-conv-new}
\lim_{t\to\infty} \langle x_k(t),\, S_{BC} x_1(0) \rangle = 0,
\qquad
\bigl|\langle x_k(t),\, S_{BC} x_1(0) \rangle\bigr|
\;\lesssim\; e^{-O(1)t}.
\eeq
and  $x_k(t)$ converges  to some unit vector $\xi \in \langle S_{BC} x_1(0) \rangle^\perp$.
\end{itemize}

Moreover, after possibly replacing each $x_i$ by its reflection $-x_i$, we may assume without loss of generality that
\[
\lim_{t\to\infty} x_i(t)=x_1(0),\qquad i\le k-1.
\]
\end{proposition}

\begin{proof} 
For simplicity of notation, we may assume $k=L$ and proceed by induction on $L$. Our main tool is Proposition \ref{conv-to-x1}. 

We first consider the case $L=2$. If $\la x_2(0),S_{BC}x_1\ra>0$, then the result follows directly from Proposition \ref{conv-to-x1}. If instead $\la x_2(0),S_{BC}x_1\ra<0$, we define
\[
\wtd x_2(t)=-x_2(t).
\]
Since 
$
\mathbf{P}_{-y}(x)=\mathbf{P}_{y}(x)
$ for all $x,y\in \S^{D-1}$, 
we obtain
\[
      \frac{d}{dt} \wtd x_2=-      \frac{d}{dt} x_2=-a_{21}\la x_2,S_{BC}x_1\ra \mathbf{P}_{x_2}(x_1)=a_{21}\la \wtd x_2,S_{BC}x_1\ra \mathbf{P}_{\wtd x_2}(x_1).
\]
Hence, by Proposition \ref{conv-to-x1} with $\la \wtd x_2(0),S_{BC}x_1\ra>0$, we deduce
\[
|\wtd x_2(t)-x_1|\lesssim e^{-O(1)t}.
\]
From this point onward, starting from the third token, we replace $(x_1,x_2)$ by $(x_1,\wtd x_2)$. We call this a positive angle rearrangement. By abuse of notation, $\wtd x_i$ may denote either $x_i$ or $-x_i$, depending on whether $x_i\to x_1$ or $x_i\to -x_1$. This completes the case $L=2$. 
Now assume we have the rearrangement $(x_1,\wtd x_2,\cdots, \wtd x_{L-1})$ such that
\[
\sum_{i=2}^{L-1} |\wtd x_i(t)-x_1|\lesssim e^{-O(1)t}.
\]

We claim that: if $\la x_L(t),S_{BC}x_1\ra\not\to 0$ as $t\to \infty$, then 
$
x_L(t)\to \pm x_1
$ as $t\to\infty$. Indeed, since $\la x_L(t),S_{BC}x_1\ra\not\to 0$ as $t\to \infty$, there exists $\eps_0>0$ such that 
either the set 
\[
\{t>0:\quad \la x_L(t),S_{BC}x_1\ra\ge \eps_0\},
\]
or
\[
\{t>0:\quad \la x_L(t),S_{BC}x_1\ra\le -\eps_0\}
\]
contains an increasing sequence of times $\{t_n\}_{n=1}^\infty$. We will show that there exists $n$ such that
\[
\la x_L(t_n),S_{BC}x_1\ra,\cdots,\la x_L(t_n),S_{BC}\wtd x_{L-1}(t_n)\ra
\]
all have the same sign (positive or negative). If this is the case, we can perform the reflection $\wtd x_L$, apply Proposition \ref{conv-to-x1}, and conclude $\wtd x_L\to x_1$ as $t\to \infty$. 

Since $\wtd x_{k}\to x_1$ for all $k\le L-1$ as $t\to\infty$, there exists $T>0$ such that for any $k\in [2,L-1]$,
\[
|\la x_L(t),S_{BC}x_1\ra-\la x_L(t),S_{BC}\wtd x_k(t)\ra|\le 0.1\eps_0\quad\text{for all}\quad t\ge T.
\]
This implies
\[
 \la x_L(t),S_{BC}x_1\ra-0.1\eps_0\le\la x_L(t),S_{BC}x_k(t)\ra\le \la x_L(t),S_{BC}x_1\ra+0.1\eps_0\quad\forall t\ge T.
\]
\begin{itemize}
    \item If there is an increasing sequence of times $\{t_n\}$ such that $\la x_L(t_n),S_{BC}x_1\ra\ge \eps_0$, then we can find $t_n\ge T$ such that
    \[
    \la x_L(t_n),S_{BC}x_k(t_n)\ra\ge 0.9\eps_0.
    \]
    Hence Proposition \ref{conv-to-x1} applied to $(x_1,\wtd x_2,\cdots,x_L)$ at time $t=t_n$ yields $x_L(t)\to x_1$ as $t\to\infty$.

    \item If there is an increasing sequence of times $\{t_n\}$ such that $\la x_L(t_n),S_{BC}x_1\ra\le - \eps_0$, then we can find $t_n\ge T$ such that
    \[
    \la x_L(t_n),S_{BC}x_k(t_n)\ra\le -0.9\eps_0.
    \]
    Hence Proposition \ref{conv-to-x1} applied to $(x_1,\wtd x_2,\cdots,\wtd x_L)$ at time $t=t_n$ yields $\wtd x_L(t)\to x_1$ as $t\to\infty$.
\end{itemize}
The claim is proved.

Now if $\la x_L(t),S_{BC}x_1\ra\to 0$, we show that
\[
\lw|\la x_L(t),S_{BC}x_1\ra
\rw|\lesssim e^{-O(1)t}.
\]
Recalling that $U_{L1}(t)=\la x_L(t),S_{BC}x_1\ra$, the above is equivalent to
\[
U_{L1}^2(t)\to 0\quad\text{as}\quad t\to\infty.
\]
Using the evolution equation of $U_{L1}$ in \ref{XU} and the fact that all previous tokens converge to $\pm x_1(0)$, there exists a time $T>0$ and positive constants $c_T,\sigma,C_{\sigma}>0$ such that
\beq\label{c1bound}
\frac{d}{dt}U_{L1}^2 \ge c_T U_{L1}^2 - C_\sigma e^{-\sigma t}\quad\text{for all}\quad t\ge T.
\eeq  
Hence for any time $T_1\ge T$, we obtain
\[
U_{L1}^2(t)\ge e^{c_0(t-T_1)}\lw[U_{L1}(T_1)^2-\frac{C_\sigma}{c_T+\sigma} e^{-\sigma {T_1}}
\rw]
+\frac{C_\sigma}{c_T+\sigma} e^{-\sigma t}, 
\qquad t \ge T_1.
\]
Since $U_{L1}^2(t)\in [0,1]$ for all $t\ge 0$, by letting $t\to\infty$ it follows that
\[
U_{L1}(T_1)^2\le \frac{C_\sigma}{c_T+\sigma}e^{-\sigma T_1}.
\]
Since the above inequality is true for any time $T_1\ge T$, we get \eqref{exp-conv-new}. Now we have
\[ \left|\dfrac{d x_k}{d t}\right|\lesssim
\big| \langle x_k(t), S_{BC} x_1(0) \rangle \big| \lesssim e^{-O(1)t}.
\]
This implies that $\{x_k(t)\}_{t>0}$ is Cauchy and hence $x_k(t)$ converges  to some unit vector $\xi \in \langle S_{BC} x_1(0) \rangle^\perp$ as $t\to\infty$. The proof is complete.

\end{proof}
\begin{remark}
Proposition~\ref{limit} shows that, starting from the third token onward, each trajectory either converges to a point in $\{\pm x_1\}$ or else its component in the $S_{BC}x_1(0)$ direction vanishes asymptotically. 
\end{remark}

In the next proposition, we establish that if all preceding tokens converge to $\{\pm x_1\}$, then the probability that the subsequent token converges to any point outside $\{\pm x_1\}$ is zero.

\begin{proposition}\label{zero-prop}
Let $L \geq 3$ and assume that $\la S_{BC}x_1(0), x_1(0) \ra > 0$.  
Denote by $K$ the set of all points $\xi \in \S^{D-1}$ such that
\[
\la \xi, S_{BC}x_1(0) \ra = 0.
\]
Then the set of initial configurations $(x_2(0), \ldots, x_L(0)) \in (\S^{D-1})^{L-1}$ for which the corresponding trajectory satisfies
\[
(x_2(t), \ldots, x_{L-1}(t), x_L(t)) \;\longrightarrow\; (\pm x_1, \ldots, \pm x_1, \xi) 
\quad \textbf{for some } \xi\in K \text{ as } t \to \infty
\]
has measure zero in $(\S^{D-1})^{L-1}$.
\end{proposition}

\begin{proof} 
Without loss of generality, we may assume that $x_k \to x_1$ for all $k \in [2,L-1]$.  
Our first goal is to show that, for each $\xi\in K$ the set of initial conditions leading to $x_L \to \xi$ has measure zero in $(\S^{D-1})^{L-1}$.  

Recall the notation
\[
X_{ij} = x_i \cdot x_j, 
\qquad 
U_{ij} = x_i \cdot S_{BC}x_j.
\]
For convenience, we define the limiting configuration associated with $\xi$ by
\[
x_\star = x_\star(\xi) := (x_1, \ldots, x_1, \xi) \in (\S^{D-1})^L.
\]
We perturb the configuration $(x_i)_{i=1}^L$ around $x_\star$ by setting
\[
\begin{cases}
\wtd x_1 = 0, \quad 
x_i = x_1 + \delta \wtd x_i + O(\delta^2)
& \text{for } 2 \leq i \leq L-1, \\[4pt]
x_L = \xi + \delta \wtd x_L + O(\delta^2), & \\[6pt]
(x_1, \ldots, x_L)
= x_\star + \delta \,(0, \wtd x_2, \ldots, \wtd x_L)
+ O(\delta^2). &
\end{cases}
\]
Here
\[
x_1\cdot\wtd x_i=0
\quad\text{for }2\leq i\leq L-1,
\qquad
\xi\cdot\wtd x_L=0.
\]
For $i,j \leq L$, we introduce the first variations
\[
\wtd X_{ij}
= \wtd x_i\cdot x_{\star,j}
  + x_{\star,i}\cdot\wtd x_j,
\qquad
\wtd U_{ij}
= \wtd x_i\cdot S_{BC}x_{\star,j}
  + x_{\star,i}\cdot S_{BC}\wtd x_j,
\]
and we set
\[
\gamma = S_{BC}x_1 \cdot x_1 > 0.
\]

We obtain the following first-order expansions:
\[
\begin{cases}
X_{Li} &= \xi \cdot x_1 
   + \delta \big[ \wtd X_{L1} + \xi \cdot \wtd x_i \big] 
   + O(\delta^2), \qquad 1 \le i \le L-1, \\[6pt]
U_{Li} &= \delta \big[ \xi \cdot S_{BC}\wtd x_i + \wtd U_{L1} \big] 
   + O(\delta^2), \qquad 1 \le i \le L-1, \\[6pt]
X_{ij} &= 1 + \delta\wtd X_{ij} + O(\delta^2)
   = 1 + O(\delta^2), \qquad 1 \le i,j \le L-1, \\[6pt]
U_{ij} &= \gamma
   + \delta\,S_{BC}x_1\cdot(\wtd x_i+\wtd x_j)
   + O(\delta^2), \qquad 1 \le i,j \le L-1.
\end{cases}
\]
Substituting the above expansions into the equations of Lemma~\ref{XU} and collecting the $O(\delta)$ terms, we obtain:

\begin{itemize}
\item[(i)] Linearized equation for $U_{L1}$:
\[
\gamma^{-1}\,\partial_t \wtd U_{L1}
= a_{L1}(x_\star)\wtd U_{L1}
+ \sum_{j=2}^{L-1}a_{Lj}(x_\star)
\big[\wtd U_{L1}+\xi\cdot S_{BC}\wtd x_j\big].
\]

\item[(ii)] For $2 \leq j \leq L-1$, we have
\[
\begin{aligned}
\gamma^{-1}\,\partial_t
\big[\xi\cdot S_{BC}\wtd x_j+\wtd U_{L1}\big]
&=
\big(a_{L1}(x_\star)+a_{j1}(x_\star)\big)\wtd U_{L1} \\
&\quad+
\Big(a_{Lj}(x_\star)-\sum_{k=1}^{j-1}a_{jk}(x_\star)\Big)
\big[\xi\cdot S_{BC}\wtd x_j+\wtd U_{L1}\big] \\
&\quad+
\sum_{k=2}^{j-1}a_{jk}(x_\star)
\big[\xi\cdot S_{BC}\wtd x_k+\wtd U_{L1}\big] \\
&\quad+
\sum_{\substack{2\leq k\leq L-1\\k\neq j}}
a_{Lk}(x_\star)
\big[\xi\cdot S_{BC}\wtd x_k+\wtd U_{L1}\big].
\end{aligned}
\]
\end{itemize}

We now introduce the vector $\vec{X} \in \R^{L-1}$ defined by
\[
\vec{X}(t) = 
\begin{bmatrix}
\wtd U_{L1} \\[4pt]
\xi\cdot S_{BC}\wtd x_2+\wtd U_{L1} \\[4pt]
\vdots \\[4pt]
\xi\cdot S_{BC}\wtd x_{L-1}+\wtd U_{L1}
\end{bmatrix}
= [X_1, X_2, \ldots, X_{L-1}]^T.
\]
The vector $\vec{X}$ collects the first variations of the interactions
$U_{L1},\ldots,U_{L,L-1}$ relative to the limiting configuration
$x_\star \in (\S^{D-1})^L$.
From the previous computations we obtain
\[
\begin{aligned}
\gamma^{-1}\,\partial_t X_1 
&= a_{L1}X_1+\sum_{j=2}^{L-1}a_{Lj}X_j, \\[6pt]
\gamma^{-1}\,\partial_t X_j 
&= (a_{L1}+a_{j1})X_1
+ \sum_{k=2}^{j-1}(a_{jk}+a_{Lk})X_k \\
&\quad+
\Big(a_{Lj}-\sum_{k=1}^{j-1}a_{jk}\Big)X_j
+ \sum_{k=j+1}^{L-1}a_{Lk}X_k,
\qquad 2\leq j\leq L-1.
\end{aligned}
\]

Hence the vector $\vec{X}$ satisfies the linear system
\[
\gamma^{-1}\,\partial_t\vec{X}(t)
=J(x_\star)\vec{X}(t),
\]
where $J(x_\star)\in\R^{(L-1)\times(L-1)}$ is the coefficient matrix
of the reduced linearized system with rows
$R_i=(J(x_\star))_{i,:}$ defined as follows
(all coefficients $a_{ij}$ are evaluated at $x_\star=x_\star(\xi)$):

\[
\begin{aligned}
R_1 &= 
\begin{bmatrix}
a_{L1} & a_{L2} & a_{L3} & \cdots & a_{L,L-1}
\end{bmatrix}, \\[6pt]
R_2 &= 
\begin{bmatrix}
a_{L1}+a_{21} & a_{L2}-a_{21} &
a_{L3} & \cdots & a_{L,L-1}
\end{bmatrix}, \\[6pt]
R_3 &= 
\begin{bmatrix}
a_{L1}+a_{31} & a_{32}+a_{L2} &
a_{L3}-(a_{31}+a_{32}) & \cdots & a_{L,L-1}
\end{bmatrix}, \\[6pt]
&\ \ \vdots \\[6pt]
R_{L-2} &= 
\begin{bmatrix}
a_{L1}+a_{L-2,1} &
a_{L-2,2}+a_{L2} &
a_{L-2,3}+a_{L3} & \cdots &
a_{L-2,L-3}+a_{L,L-3} &
a_{L,L-2}-\sum_{k=1}^{L-3}a_{L-2,k} &
a_{L,L-1}
\end{bmatrix}, \\[6pt]
R_{L-1} &= 
\begin{bmatrix}
a_{L1}+a_{L-1,1} &
a_{L-1,2}+a_{L2} & \cdots &
a_{L-1,L-2}+a_{L,L-2} &
a_{L,L-1}-\sum_{k=1}^{L-2}a_{L-1,k}
\end{bmatrix}.
\end{aligned}
\]
We now show that the matrix $J(x_\star)$ always possesses at least one positive eigenvalue.
Every row of $J(x_\star)$ has sum
$\sum_{j=1}^{L-1}a_{Lj}(x_\star)$. Hence
\[
J(x_\star)
\begin{bmatrix}
1\\ \vdots\\ 1
\end{bmatrix}
=
\left(\sum_{j=1}^{L-1}a_{Lj}(x_\star)\right)
\begin{bmatrix}
1\\ \vdots\\ 1
\end{bmatrix}.
\]
To verify that this eigenvalue corresponds to a tangent direction
of the original system, take
\[
\wtd x_i=0 \quad (1\leq i\leq L-1),
\qquad
\wtd x_L=P_\xi(x_1).
\]
Since $\xi\cdot S_{BC}x_1=0$, we have
\[
\wtd x_L\cdot S_{BC}x_1
=
P_\xi(x_1)\cdot S_{BC}x_1
=
\gamma.
\]
The earlier perturbations remain zero, and the linearized
last-token equation gives
\[
\begin{aligned}
\partial_t\wtd x_L
&=
\left(\sum_{j=1}^{L-1}a_{Lj}(x_\star)\right)
(\wtd x_L\cdot S_{BC}x_1)P_\xi(x_1)\\
&=
\gamma\left(\sum_{j=1}^{L-1}a_{Lj}(x_\star)\right)\wtd x_L.
\end{aligned}
\]
Consequently, the tangent linearization of the original system admits
the positive eigenvalue
\[
\lambda_+
=
\gamma\sum_{j=1}^{L-1}a_{Lj}(x_\star)>0.
\]
We now prove that, for almost every initial configuration 
\[
\vec{x}_0 = (x_2(0), \ldots, x_L(0)) \in (\S^{D-1})^{L-1},
\]
the trajectory converges with limits in $\{\pm x_1\}$. For $t \geq 0$, set
\[
\vec{x}(t) = (x_2(t), x_3(t), \ldots, x_L(t)) \in (\S^{D-1})^{L-1}.
\]

Define the flow map $\Phi : (\S^{D-1})^{L-1}\to (\S^{D-1})^{L-1}$ by
\[
\vec{x}_0 = (x_2,\ldots,x_L)(0) \;\mapsto\; \vec{x}(t)=(x_2,\ldots,x_L)(t).
\]
If $\vec{x}(t)=\Phi^t(\vec{x}_0)$ does not converge to $(x_1,\ldots,x_1)$ (up to a reflection of tokens), then by Proposition \ref{limit}, we have 
\[
\la x_L(t),S_{BC}x_1\ra\to 0
\]
as $t\to\infty$. 
Now fix any $\xi\in \S^{D-1}$ such that $\la \xi,S_{BC}x_1(0)\ra=0$. 
If $\Phi^t(\vec x_0)\to (\pm x_1,\cdots,\pm x_1,\xi)=x_\star(\xi)$
as $t\to\infty$, then for any $t\in\mathbb Q^+$, we have 
\begin{equation}\label{mea-zero}
\Phi^{t}(\vec{x}_0) \in W_{\text{loc}}^{\text{sc}}(x_\star(\xi)),
\end{equation}
where $W_{\text{loc}}^{\text{sc}}(x_\star(\xi))$ denotes the local center-stable manifold at $x_\star(\xi)$.  
Since the linearization at $x_\star(\xi)\in(\mathbb{S}^{d-1})^{L-1}$ admits at least one positive eigenvalue,  
\[
\lambda_+(\xi)
=
\gamma\sum_{j=1}^{L-1}a_{Lj}(x_\star(\xi))>0,
\]
in the tangent direction $(0,\ldots,0,P_\xi(x_1))$,  
the local stable--center manifold $W_{\mathrm{loc}}^{\mathrm{sc}}(x_\star(\xi))$ has dimension at most $(d-1)(L-1)-1$. Consequently, it has measure zero in $(\mathbb{S}^{d-1})^{L-1}$. 
From \eqref{mea-zero} it follows that
\[
\vec{x}_0 \in \bigcup_{t \in \mathbb Q^+} \Phi^{-t}\!\left(W_{\text{loc}}^{\text{sc}}(x_\star(\xi))\right),
\]
Hence $\vec x_0$ belongs to a set of measure zero, namely 
$N_{\xi}:=\bigcup_{t \in \mathbb Q^+} \Phi^{-t}\!\left(W_{\text{loc}}^{\text{sc}}(x_\star(\xi))\right).$
This completes the first goal of the proof.

We now note that for each $\xi\in K$, there exists an open ball
$B_{\xi}=B(x_\star(\xi), r_{\xi})$ such that if
$\vec{x}(t_1)\in B_{\xi}\setminus W_{\text{loc}}^{\text{sc}}(x_\star(\xi))$ for some $t_1>0$, then there exists $t_2>t_1$ such that $\vec{x}(t_2)\notin B_{\xi}$. As a consequence, if 
$\vec{x}_0\notin N_{\xi}$ then $\vec{x}(t)$ does not converge
to any point in $B_{\xi}$. Moreover, since $K$ is compact,
there exist  $\xi_1,...,\xi_m\in K$ such that
$$\{(x_1,...,x_1, \xi)\in (\mathbb{S}^{D-1})^{L-1}: \xi\in K\} \subset \bigcup_{j=1}^mB_{\xi_j}.$$
Hence, if $\vec{x}_0\notin N:=\bigcup_{j=1}^{m}N_{\xi_j}$ then $\vec{x}(t)$ does not converge to any point of the form 
$(x_1,...,x_1, \xi)$, $\xi\in K$.

The proof is complete.
\end{proof}
Before proving Proposition \ref{conv-to-x1}, we establish a useful lemma.

\begin{lemma}\label{son-lem} 
Let $g:[t_0,\infty)\to \R$ be a differentiable function such that $g'$ is continuous.  
Assume $g(t_0)>\delta$ and 
\[
g'(T) > 0 \quad \text{whenever } g(T)=\delta,
\]
then $g(t)\ge \delta$ for all $t \geq t_0$.
\end{lemma}

\begin{proof}
 Assume this is not true, since $g(t_0)>\delta$, there exists a smallest time $T>t_0$ such that $g(T)=\delta$. Since $g'$ is continuous, there exists $\eps_0>0$ such that 
\[
g'(t)>0\quad\text{for all}\quad t\in (T-\eps_0,T+\eps_0).
\]
This implies $g(t)<g(T)=\delta$ for all $t\in (T-\eps_0,T)$. By continuity of $g$ and the fact that $g(t_0)>\delta$, there exists $T_1\in (t_0,T)$ such that $g(T_1)=\delta$, violating the minimality of $T$. The proof is complete, and hence we have 
\[
g(t)>\delta\quad\text{for all}\quad t>t_0.
\]
\end{proof}

The following establishes Proposition \ref{conv-to-x1} in the special case $L=2$.

\begin{proposition}[Proposition \ref{conv-to-x1} for $L=2$]\label{2ndtoken} 
Assume that $\la x_1(0),S_{BC}x_1(0)\ra>0$. Then
\[
\lim_{t\to\infty} x_2(t) = x_1(0)
\quad \text{whenever} \quad 
\la x_2(0),S_{BC}x_1(0)\ra>0.
\]
Moreover, the following hold:
\begin{itemize}
\item Exponential convergence:
\[
|x_2(t)-x_1(0)| \;\le\; |x_2(0)-x_1(0)|\,e^{-O(1)\,t},
\]
\item Positivity preservation:
\[
\la x_2(t),S_{BC}x_1(0)\ra >0, 
\qquad \text{for all } t>0.
\]
\end{itemize}
\end{proposition}
\begin{proof} 
Since 
\[
|x_2(t)-x_1|^2 = 2\bigl[1-\la x_1, x_2(t)\ra\bigr],
\]
it suffices to prove that 
\[
\lim_{t\to\infty}\la x_2(t),x_1\ra = 1.
\]

From the dynamical system~\eqref{new-eq}, we have
\[
\frac{d}{dt} x_2 = \la x_2, S_{BC}x_1\ra\, a_{21}\bigl(x_1-(x_1\cdot x_2)x_2\bigr),
\]
where $a_{21}>0$ for all $t>0$.  
Define
\[
f(t)=\la x_2(t),x_1\ra, 
\qquad 
g(t)=\la x_2(t),S_{BC}x_1\ra.
\]
A direct computation yields
\[
\begin{aligned}
f' &= a_{21} g(1-f^2),\\
g' &= a_{21} g\,(x_1\cdot S_{BC}x_1 - fg).
\end{aligned}
\]
\emph{Step 1: Positivity of $g(t)$.}  
We claim that
\begin{equation}\label{ran-2}
g(t)\;\geq\;\min\{\,S_{BC}x_1\cdot x_1,\; g(0)\}, \quad\text{for all}\quad t>0.
\end{equation}
At $t=0$, we have $g(0)=\la x_2(0),S_{BC}x_1\ra>0$.  
Fix any $\delta\in(0,\min\{S_{BC}x_1\cdot x_1,g(0)\})$.  
If $g(T)=\delta$ at some time $T$, then
\[
g'(T) = a_{21}\delta\bigl(x_1\cdot S_{BC}x_1 - \delta f(T)\bigr) 
       \;\geq\; a_{21}\delta\bigl(S_{BC}x_1\cdot x_1 - \delta\bigr) >0.
\]
By Lemma~\ref{son-lem}, this implies $g(t)>\delta$ for all $t>0$, proving \eqref{ran-2}.
~\\
\emph{Step 2: Convergence of $f(t)$.}  
From \eqref{ran-3},
\[
f' = a_{21} g(1-f^2),
\]
with $a_{21},g>0$. Hence $f'(t)>0$ whenever $f(t)\in(-1,1)$, so $f$ is strictly increasing.  
Therefore, there exists $\theta = \sup_{t\ge0} f(t) \in [-1,1]$, with necessarily $\theta>-1$.  
\emph{Step 3: Exponential convergence.}  
For large $t$, we estimate
\[
\frac{d}{dt}(1-f) = -f' = -a_{21}g(1-f)(1+f)
\;\lesssim\; -(1+\theta)(1-f(t)).
\]
Integrating, we obtain
\[
1-f(t) \;\le\; (1-f(0))e^{-O(1)\,t}.
\]
Thus $f(t)\to 1$ exponentially fast as $t\to\infty$.  

Combining Steps 1–3, we conclude that $\lim_{t\to\infty}x_2(t)=x_1(0)$, which completes the proof.
\end{proof}

We are ready to provide the proof of Proposition~\ref{conv-to-x1}.

\begin{proof}[{\textbf{Proof of Proposition \ref{conv-to-x1}}}]\label{proof-12}
We proceed by induction on $L$.  
The case $L=2$ has already been established in Proposition~\ref{2ndtoken}.  
Assume now that $L \geq 3$ and that the proposition holds for $L-1$, i.e.,
\[
\sum_{i=1}^{L-1} |x_i(t)-x_1| \lesssim e^{-O(1)t},
\]
and moreover $\la x_i(t), S_{BC}x_j(t)\ra > 0$ for all $i,j=1,\ldots,L-1$.  
We will show that $x_L(t)\to x_1$ with exponential decay rate.  
The proof of Proposition~\ref{conv-to-x1} is divided into several claims below.

\paragraph{Claim 1.} \textit{For any $i,j$, we have $U_{ij}(t)=\la x_i(t),S_{BC}x_j(t)\ra>0$ for all $t>0$.}

\begin{proof}[Proof of Claim 1]
Indeed, by the induction hypothesis in Proposition~\ref{conv-to-x1}, 
it suffices to consider $U_{Lj}$ with $1\le j\le L$.  
We define the set 
\[
\mathcal S = [0,\infty)^L \subset \R^L.
\]
By the assumption of Proposition~\ref{conv-to-x1}, we have
\[
(U_{L1}(0), U_{L2}(0), \ldots, U_{LL}(0)) \in \mathcal S.
\]
We now show that the trajectory 
\[
(U_{L1}(t), U_{L2}(t), \ldots, U_{LL}(t))
\]
remains inside $\mathcal S$ for all $t>0$.  
Suppose, for contradiction, that the trajectory leaves $\mathcal S$.  
Since it is bounded within a box, it must cross the boundary of $\mathcal S$ at some finite time.  
There are two possible scenarios:  

\begin{itemize}
\item There exists a smallest exit time $T>0$ and a unique index $i_\star \in [1,L]$ such that 
\[
U_{Li_\star}(T)=0, \qquad U_{Li_\star}(t)>0 \quad \text{for all } t \in (0,T),
\] 
while 
\[
\sum_{k\neq i_\star} U_{Lk}^2(T) > 0.
\]
That is, the trajectory exits $\mathcal S$ through the $i_\star^{\text{th}}$ component $U_{Li_\star}$, rather than exiting all components simultaneously at time $T$.
\item There exists a smallest exit time $T>0$ such that $U_{Li}(T)=0$ for all $i$, with $U_{Li}(t)>0$ for all $t \in (0,T)$.  
In this case, the trajectory exits $\mathcal S$ all at once at time $T$.~\\
\end{itemize}
We now consider the first case, when the trajectory exits $\mathcal S$ only at the component $i_\star$.  
By Lemma~\ref{XU}, for $i_\star>1$ we have
\[\bega 
U'_{Li_\star}(T)
&=\sum_{k\le L-1}a_{Lk}U_{Lk}(T)
   +\sum_{k\le i_\star-1}a_{jk}U_{jk}(T)\lw[U_{Lk}(T)-X_{jk}(T)U_{Li_\star}(T)\rw]\\
&\ge \sum_{k\le L-1}a_{Lk}U_{Lk}(T)\\
&>0,
\enda 
\]
since at least one of the $U_{Lk}(T)$ remains strictly positive.  

For $i_\star=1$, we obtain
\[\bega 
U'_{L1}(T)
&=\sum_{j\le L-1} a_{Lj}U_{Lj}(T)\lw[U_{j1}(T)-X_{Lj}(T)U_{L1}(T)\rw]\\
&=\sum_{j\le L-1}a_{Lj} U_{Lj}(T)U_{j1}(T)\\
&>0.
\enda 
\]

Thus, in both cases we conclude that $U'_{Li_\star}(T)>0$.

By continuity of $U'_{Li_\star}$, there exists $\eps_0>0$ such that 
\[
U_{Li_\star}'(t)>0 \quad \text{for all } t\in (T-\eps_0,T+\eps_0).
\]
Hence $U_{Li_\star}(t)$ is strictly increasing on $(T-\eps_0,T+\eps_0)$, which implies
\[
U_{Li_\star}(t)<U_{Li_\star}(T)=0 \quad \forall\, t\in (T-\eps_0,T).
\]
Since $U_{Li_\star}(0)>0$, there must exist some $T_1\in (0,T-\eps_0)$ such that $U_{Li_\star}(T_1)=0$. 
Moreover, $U_{Li}(t)>0$ for all $t\in [0,T_1]\subset [0,T]$. 
This contradicts the minimality of $T$.  
Therefore, the first case cannot occur.

Now we consider the second case, when the trajectory 
\[
(U_{L1}(t),U_{L2}(t),\ldots, U_{LL}(t))
\]
exits $\mathcal S$ simultaneously at some time $T$. 
In this situation, Lemma~\ref{XU} implies that 
\[
U_{Lk}'(T)=0 \qquad \text{for all } k\in [1,L].
\]
Thus the trajectory passes through the equilibrium point $0\in\R^L$ at time $t=T$.  
Choose $\eps_0>0$ sufficiently small, and consider the flow ODE for $(U_{L1}(t),U_{L2}(t),\ldots, U_{LL}(t))$ on the interval $I=(T-\eps_0,T+\eps_0)$. On this interval we observe two distinct flows satisfying the same system with the same initial data at $t=T$:  
the trivial solution $(U_{L1}(t),\ldots, U_{LL}(t))\equiv 0$ and the trajectory $(U_{L1}(t),\ldots, U_{LL}(t))$ with $U_{Li}(t)>0$ for all $t\in (T-\eps_0,T)$.  
This contradicts the uniqueness of solutions to the differential equations on $I$.  

Hence the trajectory cannot exit the boundary of $\mathcal S$ simultaneously at all components.  
This completes the proof for the second case and concludes the proof of Claim~1.

\end{proof}
\paragraph{Claim 2.}\label{UL1} \textit{There exist a time $T > 0$ and a constant $c_T > 0$ such that $U_{ij}(t) \ge c_T$ for all $t \ge T$ and all $1 \le i,j \le L$.}

\begin{proof}[Proof of Claim 2]
For $i,j \le L-1$, we observe that
\[
\lim_{t \to \infty} U_{ij}(t) = \lim_{t \to \infty} \langle x_i(t), S_{BC} x_j(t) \rangle = \langle x_1, S_{BC} x_1 \rangle > 0.
\]
Therefore, the claim holds for such $i$ and $j$. It remains to consider the case of $U_{Lj}(t)$ for $1 \le j \le L$.
Now, for any $i,j\le L-1$, there exists a time $T>0$ such that
\[
U_{ij}(t)\ge 0.9\, (S_{BC}x_1\cdot x_1), \quad \forall t\ge T.
\]
We aim to show that for $2\le j\le L$,
\[
U_{Lj}(t) \ge \min\Big\{ 0.8\, (S_{BC}x_1\cdot x_1),\, U_{L1}(T), \ldots, U_{Lj}(T) \Big\}, \quad \forall t\ge T.
\]
This will be established by induction on $j$.  

For the base case $j=1$, using Lemma~\ref{XU}, we have
\[
U'_{L1}(t) = \sum_{k\le L-1} a_{Lk} U_{Lk} \big[ U_{k1} - X_{Lk} U_{L1} \big].
\]
Choose any $\delta>0$ satisfying
\[
\delta < \min \big\{ 0.8\, (S_{BC}x_1\cdot x_1),\, U_{L1}(T) \big\}.
\]
We claim that
\beq\label{ran-3}
U_{L1}(t) \ge \delta, \quad \forall t \ge T.
\eeq
Indeed, by Lemma~\ref{son-lem}, it suffices to check that whenever $U_{L1}(T_\star)=\delta$, we have
\[\bega
U'_{L1}(T_\star) &= \sum_{k\le L-1} a_{Lk} U_{Lk}(T_\star) \big[ U_{k1}(T_\star) - X_{Lk}(T_\star)\, \delta \big] \\
&\ge \sum_{k\le L-1} a_{Lk} U_{Lk}(T_\star) \big[ 0.9\, (S_{BC}x_1\cdot x_1) - \delta \big] > 0.
\enda
\]
This establishes \eqref{ran-3} for $j=1$.
Now, assume that for all $k \le j-1$,
\[
\inf_{t \ge T} U_{Lk}(t) \ge \min\left\{ 0.8\, \langle S_{BC}x_1, x_1 \rangle,\, U_{L1}(T), \ldots, U_{Lk}(T) \right\}.
\]
We will show that
\[
\inf_{t \ge T} U_{Lj}(t) \ge \min\left\{ 0.8\, \langle S_{BC}x_1, x_1 \rangle,\, U_{L1}(T), \ldots, U_{Lj}(T) \right\}.
\]
To this end, fix any $\delta$ satisfying
\[
0 < \delta < \min\left\{ 0.8\, \langle S_{BC}x_1, x_1 \rangle,\, U_{L1}(T), \ldots, U_{Lj}(T) \right\}.
\]
We will prove that $U_{Lj}(t) \ge \delta$ for all $t \ge T$.

Recall from Lemma~\ref{XU} that
\[
\begin{aligned}
U'_{Lj} = &\sum_{k \le L-1} a_{Lk} U_{Lk} \left[ U_{kj} - X_{Lk} U_{Lj} \right] 
          + \sum_{k \le j-1} a_{jk} U_{jk} \left[ U_{Lk} - X_{jk} U_{Lj} \right].
\end{aligned}
\]
Suppose $U_{Lj}(T_\star) = \delta$ for some $T_\star\ge T$. Then, we have
\[
\begin{aligned}
U'_{Lj}(T_\star) \ge &\sum_{k \le L-1} a_{Lk} U_{Lk}(T_\star) \left[ U_{kj}(T_\star) - \delta \right] 
              + \sum_{k \le j-1} a_{jk} U_{jk}(T_\star) \left[ U_{Lk}(T_\star) - \delta \right] \\
           \ge &\sum_{k \le j-1} a_{jk} U_{jk} \left[ U_{Lk}(T_\star) - \delta \right] 
           > 0.
\end{aligned}
\]
The claim then follows by an application of Lemma~\ref{son-lem}.
\end{proof}

\paragraph{Claim 3.} \textit{There exist a time $T > 0$ and a constant $c_T > 0$ such that
\begin{equation}\label{ran-5}
1 + X_{L1}(t) \ge c_T \quad \text{for all } t \ge T.
\end{equation}
}
\begin{proof}[Proof of Claim 3]
From Lemma~\ref{XU}, we have
\[
\begin{aligned}
X_{L1}'(t)
&= a_{L1} U_{L1} (1 - X_{L1}^2) \\
&\quad + \sum_{j=2}^{L-1} a_{Lj} U_{Lj} \left[ X_{j1} - X_{Lj} X_{L1} \right].
\end{aligned}
\]

Fix $\varepsilon \in (0,1)$ sufficiently small. Since $X_{j1}(t) \to 1$ as $t \to \infty$ for all $j \le L-1$, there exists $T > 0$ such that
\[
X_{j1}(t) > 1 - \varepsilon \quad \text{for all } t \ge T.
\]

If $X_{L1}(t) \to 1$ as $t \to \infty$, then the claim holds trivially. Suppose, to the contrary, that $X_{L1}(t) \not\to 1$ as $t \to \infty$. Then there exists an increasing sequence $\{t_n\}_{n=1}^\infty \subset [T, \infty)$ such that
\[
X_{L1}(t_n) \le 1 - \varepsilon.
\]

We now show that not all $t_n$ can satisfy $X_{L1}(t_n) = -1$. Assume for contradiction that
\[
X_{L1}(t_n) = \langle x_L(t_n), x_1 \rangle = -1 \quad \text{for all } n \ge 1.
\]
This implies
$
x_L(t_n) = -x_1 
$ for all $n\ge 1$.
Consequently,
\[
U_{L1}(t_n) = \langle x_L(t_n), S_{BC} x_1 \rangle = -\langle x_1, S_{BC} x_1 \rangle < 0,
\]
which contradicts the fact that $U_{L1}(t) > 0$ for all $t > 0$.

Therefore, there exists a time $t_\star \ge T$ such that
\[
X_{L1}(t_\star) \le 1 - \varepsilon, \quad\text{and}\quad 
1 + X_{L1}(t_\star) > 0.
\]

We now prove that
\[
X_{L1}(t) \ge \min\left\{ X_{L1}(t_\star),\, 1 - \varepsilon \right\} \quad \text{for all } t \ge t_\star,
\]
which will imply inequality~\eqref{ran-5}.

To this end, fix any $\delta \in (-1, \min\left\{ X_{L1}(t_\star),\, 1 - \varepsilon \right\})$. We will show that $X_{L1}(t) \ge \delta$ for all $t \ge t_\star$ by applying Lemma~\ref{son-lem}.

Indeed, suppose $X_{L1}(t) = \delta$ for some $t \ge t_\star$. Then, using the earlier lower bounds and the fact that $U_{ij}(t) > 0$, we obtain
\[
\begin{aligned}
X_{L1}'(t)
&\ge a_{L1} U_{L1} (1 - \delta^2) + \sum_{j=2}^{L-1} a_{Lj} U_{Lj} \left[ (1 - \varepsilon) - \delta \right] > 0.
\end{aligned}
\]
This verifies the condition of Lemma~\ref{son-lem}, completing the proof of Claim 3.
\end{proof}

\paragraph{Claim 4.} \textit{There exist a time $T > 0$ and a constant $c_T > 0$ such that for all $t \ge T$,
\[
1 - X_{L1}(t) \lesssim e^{-c_T t}.
\]}

\begin{proof}[Proof of Claim 4]
By the preceding claims and the induction hypothesis, there exist $T > 0$ and $c_T > 0$ such that for all $t \ge T$,
\[
1 + X_{L1}(t) \ge c_T, \quad U_{ij}(t) \ge c_T, \quad \max_{2 \le j \le L-1} (1 - X_{j1}(t)) \lesssim e^{-c_T t}.
\]

For $t \ge T$, we compute
\[
\begin{aligned}
\frac{d}{dt} \left[ 1 - X_{L1}(t) \right]
&= -X'_{L1}(t)= -a_{L1} U_{L1} (1 - X_{L1}^2) - \sum_{j=2}^{L-1} a_{Lj} U_{Lj} \left[ X_{j1} - X_{Lj} X_{L1} \right] \\
&= -a_{L1} U_{L1} (1 - X_{L1}^2) - \sum_{j=2}^{L-1} a_{Lj} U_{Lj} (X_{j1} - 1)- \sum_{j=2}^{L-1} a_{Lj} U_{Lj} (1 - X_{Lj} X_{L1}) \\
&= -a_{L1} U_{L1} (1 - X_{L1})(1 + X_{L1}) + \sum_{j=2}^{L-1} a_{Lj} U_{Lj} (1 - X_{j1})  - \sum_{j=2}^{L-1} a_{Lj} U_{Lj} (1 - X_{Lj} X_{L1}).
\end{aligned}
\]
Note that the third term is non-positive, since $|X_{Lj} X_{L1}| \le 1$, so we may drop it to obtain an upper bound:
\[
\begin{aligned}
\frac{d}{dt} \left[ 1 - X_{L1}(t) \right]
&\le -a_{L1} U_{L1} (1 + X_{L1})(1 - X_{L1}) + \sum_{j=2}^{L-1} a_{Lj} U_{Lj} (1 - X_{j1}) \\
&\le -c_1 (1 - X_{L1}(t)) + C_0 e^{-c_T t},
\end{aligned}
\]
for some positive constants $c_1, C_0 > 0$, where we have used the lower bounds $U_{L1}(t) \ge c_T$, $1 + X_{L1}(t) \ge c_T$, and the exponential decay $1 - X_{j1}(t) \lesssim e^{-c_T t}$.

Applying the Gronwall inequality to this differential inequality yields
\[
1 - X_{L1}(t) \lesssim e^{-\min\{c_1, c_T\} t},
\]
which completes the proof of Claim 4.
\end{proof}

Proposition~\ref{conv-to-x1} now follows from Claims 1--4.
\end{proof}

We end this subsection by the proof of Theorem \ref{all-conv}.
\begin{proof}[{\textbf{Proof of Theorem \ref{all-conv}}}]
By Proposition~\ref{2ndtoken} and Proposition~\ref{limit}, the second token converges to $\pm x_1(0)$. Moreover, by Proposition~\ref{limit}, there exists a minimal integer $k \in [3, L]$ such that, after possibly replacing each $x_i$ by its reflection $\widetilde{x}_i = -x_i$ for $i \le k-1$, one of the following two cases holds:
\begin{itemize}
    \item[(a)] $\displaystyle \sum_{i=2}^{k-1} |x_i(t) - x_1(0)| + \big| \langle x_k(t), S_{BC} x_1(0) \rangle \big| \lesssim e^{-O(1)t}$,
    \item[(b)] $\displaystyle \sum_{i=2}^{k-1} |x_i(t) - x_1(0)| + |x_k(t) - x_1(0)| \lesssim e^{-O(1)t}$.
\end{itemize}
If the case (a) holds for some $3\leq k\leq L$ then we have
\[ \left|\dfrac{d x_k}{d t}\right|\lesssim
\big| \langle x_k(t), S_{BC} x_1(0) \rangle \big| \lesssim e^{-O(1)t}.
\]
This implies that $x_k(t)$ converges  to some unit vector $\xi \in \langle S_{BC} x_1(0) \rangle^\perp$. By Proposition~\ref{zero-prop}, the set $N_k$ of initial configurations for which such a limit $\xi$ occurs is of measure zero in the phase space $(\mathbb{S}^{D-1})^{k-1}$. Denote
$$N=\bigcap_{k=3}^LN_k\times (\mathbb{S}^{D-1})^{L-k}.$$
We have $N$ is a set of measure zero in $(\mathbb{S}^{D-1})^{L-1}$. Moreover, the case (b) holds for $k=L$ if $(x_2(0),...,x_L(0)\notin N$.

This completes the proof.
\end{proof}

\subsection{A Generalization to the Time-dependent Parameters Setting}\label{sec:time-dependence_appendix}
In this section, we consider the problem \eqref{eq:new-eq_main} in the case where $S_{BC}, \alpha$ and $a$ are time-dependent. Moreover, we assume that $S_{BC}(t), \alpha(t)$ and $a(t)$ converge exponentially to $S_{\infty}$, $\alpha_{\infty}$ and
$a_{\infty}$, respectively. As a sequence, there exists 
a constant $c_0>0$ such that the functions
\begin{center}
    $\tilde{S}(s)=\begin{cases}
        S_{BC}\left(-\frac{\ln |s|}{c_0}\right), s\neq 0,\\
        S_{\infty},\qquad s=0,
    \end{cases}
    \qquad
    \tilde{\alpha}(s)=\begin{cases}
        \alpha\left(-\frac{\ln |s|}{c_0}\right), s\neq 0,\\
        \alpha_{\infty},\qquad s=0,
    \end{cases}$
    and $\quad\tilde{a}(s)=\begin{cases}
        a\left(-\frac{\ln |s|}{c}\right), s\neq 0,\\
        a_{\infty},\qquad s=0,
    \end{cases}$,
\end{center}
are differentiable on $\mathbb{R}$. In addition, the gradients of these functions vanish at $t=0$.

For every $1\leq j<l\leq L$, $x=(x_1,...,x_L)\in (\mathbb{S}^{D-1})^L$ and $s\in\mathbb{R}$, we denote
$$ \tilde{A}_{lj}(s, x)= x_l^{\top} \tilde{S}(s)x_j \cdot \tilde{\triangle}_s(x_l)e^{-\tilde{a}(s) \sum_{k=j+1}^l \tilde{\triangle}_s(x_k)},$$
where
\(\tilde{\triangle}_s(x_l)=\ln(1+e^{\tilde{\alpha}(s) \cdot x_l})\).

By the same argument as for Theorem \ref{thm:all-conv_main}, we obtain the following result, whose proof we omit.
\begin{theorem}\label{thm:nonautonomous1} Consider the problem
\begin{equation}
    \begin{cases}\label{eq: nonautonomous1}
        \frac{d}{dt}x_0(t)=-c_0x_0(t),\\
    \frac{d}{dt} x_l(t)=\sum_{j=1}^{l-1}\tilde{A}_{lj}(x_0(t), \mathbf{x}(t)) \cdot \mathbf{P}_{x_l(t)}(x_j(t)), \quad l=1,\ldots,L,\\
    x_0(0)>0, \mathbf{x}(0)=(x_1(0),\ldots,x_L(0)) \in (\mathbb{S}^{D-1})^L.
    \end{cases}
\end{equation}
Let $(x_0(t), \mathbf{x}(t)= (x_1(t),\ldots,x_L(t))$ be a solution of this problem. Assume that $\tilde{S}(t)$ is symmetric for every $t$ and $S_{\infty}$ is positive definite.
Then, there exists a zero-measurable subset $N_0$ of $(0, \infty)$
satisfying the following property: if $x_0(0)\in (0, \infty)\setminus N_0$ and $x_1(0)\in \mathbb{S}^{D-1}$ then
there exists a  zero-measurable subset
$\Omega$ of $(\mathbb{S}^{D-1})^{L-1}$ such that if the initial point $(x_2(0),...,x_L(0))$ does not belong in $\Omega$  then all tokens converge exponentially to $\pm x_1(0)$.
\end{theorem}
For every $1\leq j<l\leq L$, $x=(x_1,...,x_L)\in (\mathbb{S}^{D-1})^L$ and $t\geq 0$, we denote
$$ A_{lj}(t, x)= x_l^{\top} S_{BC}(t)x_j \cdot \triangle_t(x_l)e^{-a(t) \sum_{k=j+1}^l \triangle_t(x_k)},$$
where
\(\triangle_t(x_l)=\ln(1+e^{\alpha(t) \cdot x_l})\).
Note that if $\frac{d}{dt}x_0(t)=-c_0x_0(t)$ then $x_0(t)=c\, e^{-c_0t}$ with $c=x_0(0)>0$, and it follows that
$$\tilde{A}_{lj}(x_0(t), x)=A_{lj}(t-\frac{\ln c}{c_0}, x).$$
 The following result is a direct consequence of Theorem \ref{thm:nonautonomous1}.
\begin{theorem}\label{thm:nonautonomous2} For each constant $c\in\mathbb{R}$, we consider the following problem:
\begin{equation}
    \begin{cases}\label{eq: nonautonomous2}
    \frac{d}{dt} x_l(t)=\sum_{j=1}^{l-1}A_{lj}(t+c, \mathbf{x}(t)) \cdot \mathbf{P}_{x_l(t)}(x_j(t)), \quad l=1,\ldots,L,\\
    \mathbf{x}(0)=(x_1(0),\ldots,x_L(0)) \in (\mathbb{S}^{D-1})^L.
    \end{cases}
\end{equation}
Let $(\mathbf{x}(t)= (x_1(t),\ldots,x_L(t))$ be a solution of this problem. Assume that $S_{BC}(t)$ is symmetric for every $t$ and $S_{\infty}$ is positive definite.
Then, there exists a zero-measurable subset $N_0$ of $(0, \infty)$
satisfying the following property: if $c\in \mathbb{R}\setminus N_0$ and $x_1(0)\in \mathbb{S}^{D-1}$ then
there exists a  zero-measurable subset
$\Omega$ of $(\mathbb{S}^{D-1})^{L-1}$ such that if the initial point $(x_2(0),...,x_L(0))$ does not belong in $\Omega$  then all tokens converge exponentially to $\pm x_1(0)$.
\end{theorem}

\section{Experiment - Additional Details}
\label{sec:attention-sink-appendix}
% \label{sec:experiment_appendix}

% \subsection{Predict Attention Sink}
% \label{sec:attention-sink-appendix}

\subsection{Experimental Setup}
We adopt the threshold-based metric of \cite{xiao2024efficient}, adapted to the hidden attention structure of SSMs. 
For a given layer $\ell \in \{1,\dots,L\}$ and channel $c$, let $A^{\ell,c} \in \mathbb{R}^{T \times T}$ denote the hidden attention matrix, where $T$ is the sequence length. 
The importance of token $k$ in $(\ell,c)$ is defined as
\begin{equation}
    \alpha^{\,\ell,c}_k \;=\; \frac{1}{T-k+1}\sum_{i=k}^{T} A^{\ell,c}_{i,k}.
\end{equation}
We say that channel $(\ell,c)$ exhibits an attention sink at the first token if
\begin{equation}
    \alpha^{\,\ell,c}_1 > \varepsilon,
\end{equation}
where $\varepsilon$ is a fixed threshold (we set $\varepsilon = 0.3$ in our experiments).  

The \emph{layer-wise sink rate} is given by
\begin{equation}\label{eq:sink_l}
    \mathrm{Sink}_\ell \;=\; \frac{1}{H_s}\sum_{c \in \mathcal{C}_\ell} \mathbf{1}\!\left(\alpha^{\,\ell,c}_1 > \varepsilon\right),
\end{equation}
where $\mathcal{C}_\ell$ is a subset of channels sampled from layer $\ell$, and $H_s = |\mathcal{C}_\ell|$.  
Finally, the \emph{model-level sink score} is
\begin{equation}\label{eq:sink}
    \mathrm{Sink} \;=\; \frac{1}{L}\sum_{\ell=1}^{L} \mathrm{Sink}_\ell.
\end{equation}

\subsection{Evaluation}
% \paragraph{Evaluation.} 
To empirically estimate these quantities, we evaluate the Falcon-Mamba-7B model \cite{zuo2024falcon} on the dataset of \cite{xiao2024efficient}, which consists of $170$ samples of length $T=16$. 
Since each layer contains $4096$ channels, we uniformly sample $H_s = 200$ channels per layer to estimate $\mathrm{Sink}\ell$ and the overall $\mathrm{Sink}$. We then report $\mathrm{Sink}\ell$ as a function of layer depth and the aggregate $\mathrm{Sink}$ across tokens in Figure~\ref{fig:attn-sink-analysis}.
For visualization, we additionally average the softmax-normalized attention matrices $\mathrm{softmax}(A^{\ell,c})$ across the samples to obtain per-layer attention maps, as shown in Figure~\ref{fig:attn-sink}.

\end{document}